# University of Southampton

Faculty of Engineering and Physical Sciences

Electronics and Computer Science

## Machine Learning for Flow Cytometry Data Analysis

by

**Yanhua Xu**

September 2019

Supervisor: Dr Nicolas Green

Second Examiner: Professor Elena Simperl

A dissertation submitted in partial fulfilment of the degree

of MSc Artificial Intelligence

# Abstract

Flow cytometry mainly used for detecting the characteristics of a number of biochemical substances based on the expression of specific markers in cells. It is particularly useful for detecting membrane surface receptors, antigens, ions, or during DNA/RNA expression. Not only can it be employed as a biomedical research tool for recognising distinctive types of cells in mixed populations, but it can also be used as a diagnostic tool for classifying abnormal cell populations connected with disease.

Modern flow cytometers can rapidly analyse tens of thousands of cells at the same time while also measuring multiple parameters from a single cell. However, the rapid development of flow cytometers makes it challenging for conventional analysis methods to interpret flow cytometry data. Researchers need to be able to distinguish interesting-looking cell populations manually in multi-dimensional data collected from millions of cells. Thus, it is essential to find a robust approach for analysing flow cytometry data automatically, specifically in identifying cell populations automatically.

This thesis mainly concerns discover the potential shortcoming of current automated-gating algorithms in both real datasets and synthetic datasets. Three representative automated clustering algorithms are selected to be applied, compared and evaluated by completely and partially automated gating. A subspace clustering ProClus also implemented in this thesis. The performance of ProClus in flow cytometry is not well, but it is still a useful algorithm to detect noise.

# Acknowledgements

I would first like to thank my thesis supervisor Dr Nicolas Green for his careful guidance and patience. The door to Dr Nicolas Green office was always open whenever I ran into a trouble spot or had a question about my research or writing. He consistently allowed this paper to be my own work, but steered me in the right the direction whenever he thought I needed it.

I would also like to acknowledge Profession Elena Simperl as the second examiner of this thesis, and I am gratefully indebted to her for her very valuable comments on this thesis.

Many thanks,
Yanhua Xu

**Statement of Originality**

- I have read and understood the ECS Academic Integrity information and the University's Academic Integrity Guidance for Students.
- I am aware that failure to act in accordance with the Regulations Governing Academic Integrity may lead to the imposition of penalties which, for the most serious cases, may include termination of programme.
- I consent to the University copying and distributing any or all of my work in any form and using third parties (who may be based outside the EU/EEA) to verify whether my work contains plagiarised material, and for quality assurance purposes.

***You must change the statements in the boxes if you do not agree with them.***

We expect you to acknowledge all sources of information (e.g. ideas, algorithms, data) using citations. You must also put quotation marks around any sections of text that you have copied without paraphrasing. If any figures or tables have been taken or modified from another source, you must explain this in the caption *and* cite the original source.

> **I have acknowledged all sources, and identified any content taken from elsewhere.**

If you have used any code (e.g. open-source code), reference designs, or similar resources that have been produced by anyone else, you must list them in the box below. In the report, you must explain what was used and how it relates to the work you have done.

> **I have not used any resources produced by anyone else.**

You can consult with module teaching staff/demonstrators, but you should not show anyone else your work (this includes uploading your work to publicly-accessible repositories e.g. Github, unless expressly permitted by the module leader), or help them to do theirs. For individual assignments, we expect you to work on your own. For group assignments, we expect that you work only with your allocated group. You must get permission in writing from the module teaching staff before you seek outside assistance, e.g. a proofreading service, and declare it here.

> **I did all the work myself, or with my allocated group, and have not helped anyone else.**

We expect that you have not fabricated, modified or distorted any data, evidence, references, experimental results, or other material used or presented in the report. You must clearly describe your experiments and how the results were obtained, and include all data, source code and/or designs (either in the report, or submitted as a separate file) so that your results could be reproduced.

> **The material in the report is genuine, and I have included all my data/code/designs.**

We expect that you have not previously submitted any part of this work for another assessment. You must get permission in writing from the module teaching staff before re-using any of your previously submitted work for this assessment.

> **I have not submitted any part of this work for another assessment.**

If your work involved research/studies (including surveys) on human participants, their cells or data, or on animals, you must have been granted ethical approval before the work was carried out, and any experiments must have followed these requirements. You must give details of this in the report, and list the ethical approval reference number(s) in the box below.

> **My work did not involve human participants, their cells or data, or animals.**

*ECS Statement of Originality Template, updated August 2018, Alex Weddell aiofficer@ecs.soton.ac.uk*

# Table Contents

# List of Figures

# List of Tables

# 1 Introduction

## 1.1 Aims and Objectives

Flow cytometry is a quantitative analysis technique that detects the characteristics of a number of biochemical substances based on the expression of specific markers in cells. It is particularly useful for detecting membrane surface receptors, antigens, ions, or during DNA/RNA expression. Not only can it be employed as a biomedical research tool for recognising distinctive types of cells in mixed populations, but it can also be used as a diagnostic tool for classifying abnormal cell populations connected with disease.

Modern flow cytometers can rapidly analyse tens of thousands of cells at the same time while also measuring multiple parameters from a single cell. However, the rapid development of flow cytometers makes it challenging for conventional analysis methods to interpret flow cytometry data. Researchers need to be able to distinguish interesting-looking cell populations manually in multi-dimensional data collected from millions of cells. Thus, it is essential to find a robust approach for analysing flow cytometry data automatically, specifically in identifying cell populations automatically.

Most current automated gating algorithms are based on unsupervised learning. This thesis will first compare the performance of three representative automated gating algorithms in both real datasets and synthetic datasets. Then, an algorithm that has not been used before in the field of flow cytometry will be applied. Finally, a comparative analysis of all algorithms used in this thesis will be conducted.

# 2 Background Research

## 2.1 Flow cytometry

### 2.1.1 Principles

In simple terms, flow cytometry is when a laser beam passes through a suspension of single cells. The characteristic light signals produced are then collected and analysed by detectors. In more detail, if the cells are labelled using fluorescent dyes, they can absorb the light specific wavelengths (a process called excitation) and emit light at a longer wavelength (a process called emission). Alternatively, the cells can simply scatter the light. These scattered and emitted light signals are captured by photodetectors arranged both along the line of sight direction and perpendicular to it. Following detection, the signals are digitised and stored on computers as multi-dimensional data containing embedded information about physicochemical properties of each cell. The cells' physical characteristics (size, granularity, etc.) are determined from the forward-scatter (FSC) and side-scatter signals (SSC); the chemical properties of cells are identified from the fluorescent signals. FSC and SSC are always measured, whereas the fluorescent signals are only measured in some experiments.

Flow cytometry cannot, however, measure detailed intracellular morphology, the spatial distribution of the fluorescent component, nor the component location within the cell. This is because flow cytometry can only measure cells in suspension not in solid tissues.

## 2.1.2 Parameters

Flow cytometry has three parameters: forward-scatter light (FSC), side-scatter light (SSC) and fluorescence. In a flow cytometer, the physical characteristic of a cell such as size, granularity/complexity, can be described as a detectable light signal, as Figure 1 shows.

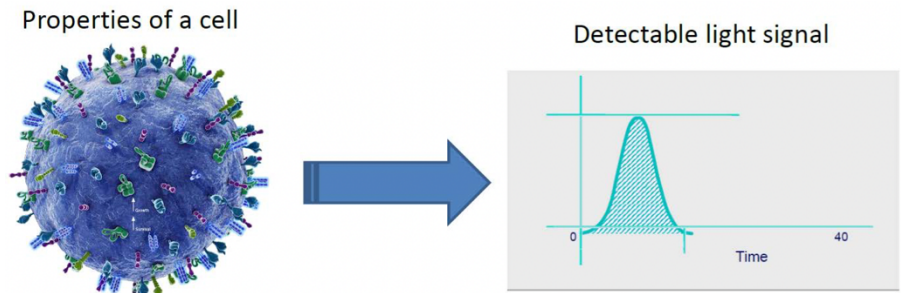

Figure 1

### 2.1.2.1 Forward Scatter (FSC) and Side Scatter (SSC)

Cells differ in size and granularity, such as lymphocytes, which are small in size and have low granularity, and monocytes, which are large but also have low granularity (Figure 2 a). Cells can be analysed based on their morphology, which requires the application of FSC and SSC (Figure 2 b)

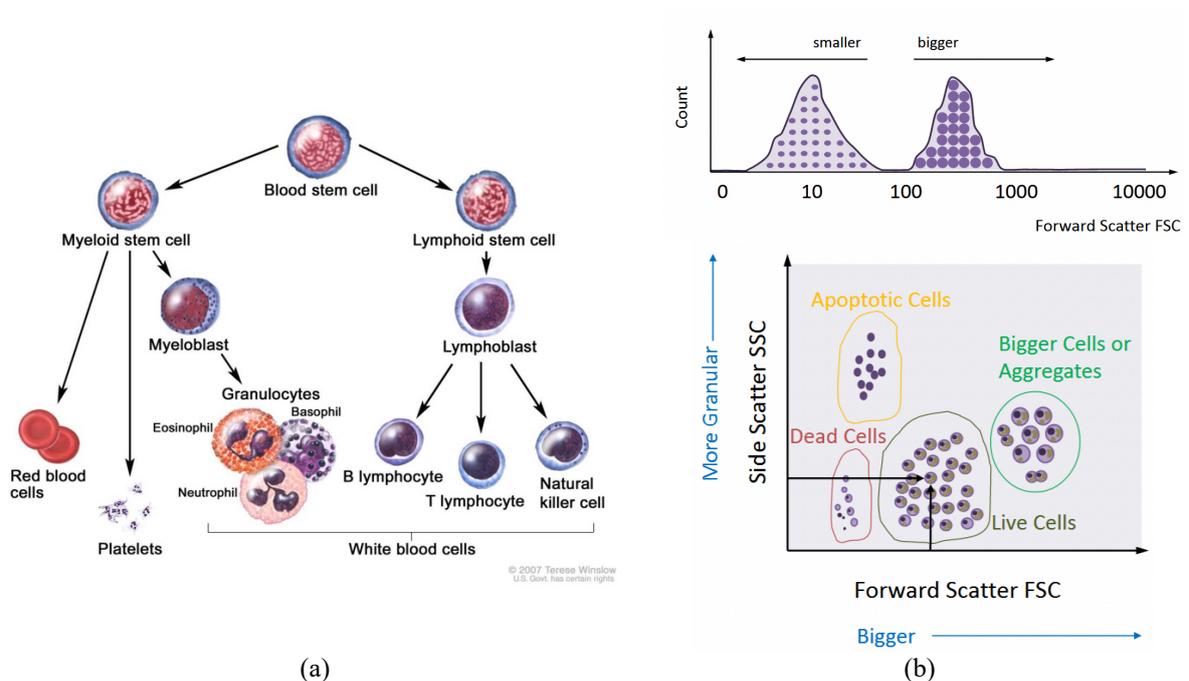

(a)            (b)

Figure 2. Analysis based on cell morphology

Forward-scatter light (FSC) is refracted by the cell along the line of the light path and is approximately proportionate to the size of the (round) cell or the area of the cell surface [1]. Larger cells may have more FSC and an enhanced FSC signal. SSC is measured by the vertical direction of the light path and concerns the complexity or the internal granularity of the cell. High-granularity cells will produce higher SSC signals (Figure 3).

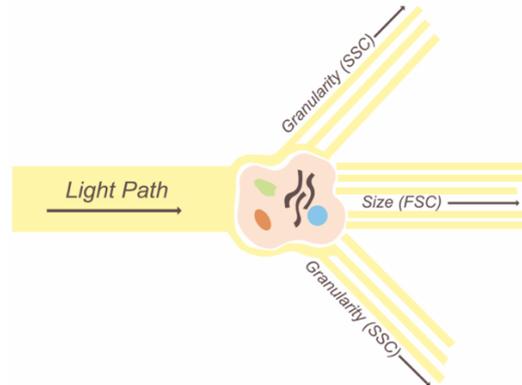

Figure 3. Detecting physical properties of the cells by using laser light [2].

### 2.1.2.2 Fluorescence

More than one laser is focused on the sample in flow cytometry. Thus, cells can be measured based on their physical properties and on their fluorescence signal intensity. Fluorescent antibodies are used to obtain information regarding the chemical properties of cells. Excited fluorescent antibodies emit fluorescence that is measured in the same direction as the side-scattered light (SSC), but the wavelength of fluorescence is longer than the wavelength of SSC. The fluorescence signal is used for detecting specific cell populations and can reflect the abundance of expressed antigens on the cell surface.

Figure 1 shows the electron changes of fluorescent antibodies. When fluorescent antibodies absorb light, its electrons are raised to a maxima energy level (this process is called excitation). After 1 to 10 nanoseconds, energy will lose in some extent, and the electron falls to a relatively lower energy level. The electron then gradually falls back to the ground state and emits light that typically has less energy than the excited state; this means that the emission wavelength is longer than the excitation wavelength. The difference between the emission wavelength and excitation wavelength is called the Stokes shift.

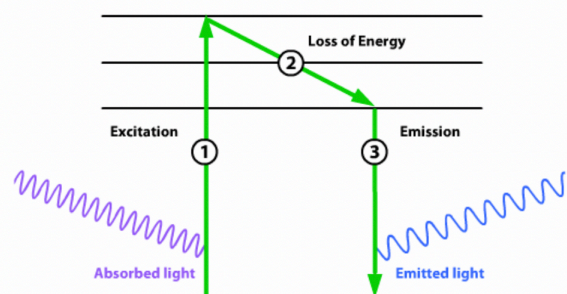

① Excitation: absorbing a photon raises an electron up to a higher energy level; ② Excited state lifetime: loss of energy by vibration, rotation; ③Emission: the electron falls back to the ground state and emits a photon with less energy than the absorbed one
Figure 4. Fluorescence electronic state diagram

## 2.1.3 Basic Components of Flow Cytometer

The components of a traditional flow cytometer are shown in Figure 5. Three main components are introduced in this section: the fluidics system, optical system and electronics system. The cells are introduced and focused in fluidics system; the light signals of the cells are generated and collected by optical system, and these are then converted to electronic signals in electronics system.

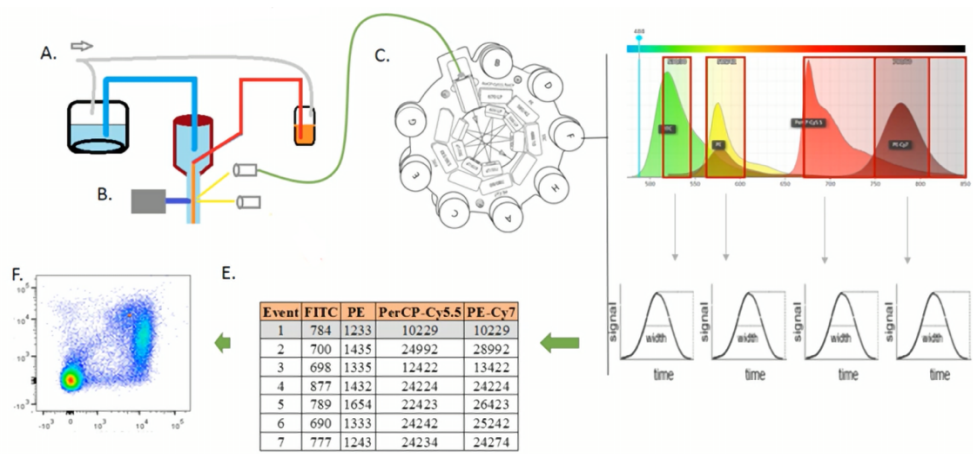

Figure 5. Component of a traditional flow cytometer. A: Fluidics; B: Laser source; C: Optics System; D-F: Electronics System

### 2.1.3.1 Fluidics

The fluidic system consists of two closely connected, yet independent, flows known as the sheath fluid flow and the sample flow. The core of the fluidics is the flow cell (or flow chamber), in which the liquid stream from a large area to a constrained channel [3]. The basic structure of the flow cell is shown in Figure 6.

Sample flow is injected into the centre of the sheath stream through a sample injection tube (SIT) in the centre of the flow cell. Sheath stream is the isosmotic solution that ensures cells within it will not die due to hypertonic or hypotonic nature of the solution. Hydrodynamic focusing aligns cells in a liquid stream. If two fluids differ enough in density and/or velocity, they do not mix, forming a two-layer stable flow (laminar flow). The sample flow is in the centre of the laminar flow; thus, the diameter of sheath flow is larger than the diameter of sample flow. The cells are then accelerated and individually pass through a laser beam for interrogation.

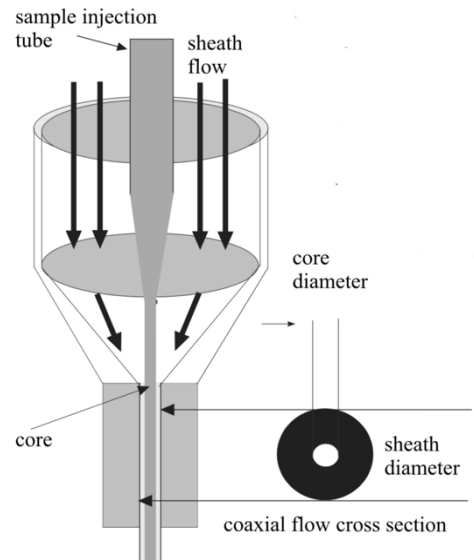

Figure 6. Basic structure of flow cell [3].

### 2.1.3.2 Optics

The heart of the flow cytometer is the optical signal, which includes the scattered light signal and the fluorescent signal. Flow cytometry analysis is based on the light signal received by the laser after irradiating the cell. To better understand flow cytometry, it is necessary to understand the optical system.

The optical system begins with the laser. The laser is one of the essential components of a flow cytometer, with each type of flow cytometer requiring at least one laser. The light signals generated and emitted from the cells that have been irradiating, are accepted by different channels through different optics.

When the laser illuminates cells in a sample flow, it produces scattered light. If the cells are bound to fluorescein, and the fluorescein is excited by the laser of this wavelength, the fluorescein will emit fluorescence. The flow cytometer collects an optical signal that includes both scattered light signals and fluorescent signals. The scattered light signal includes forward scatter (FSC) and side scatter (SSC). SSC and fluorescent signals are received from the same direction and almost mix together. Thus, flow cytometry needs to pass through the optical system to separate, according to the wavelength, the SSC from the fluorescent signal; receive them by different channels; and then indirectly reflect, according to the strength of the signal, the physicochemical properties of the cell.

The optical system consists of a series of lenses, filters, and mirror that separate the various optical signals depending on the wavelength, and the filters are the most important. In terms of their functions, the filters can be divided into three types: long pass filter, short pass filter and band pass filter, as shown in Figure 7. The optical system uses different combinations of filters to separate the optical signals. The light energy are then collected by detectors.

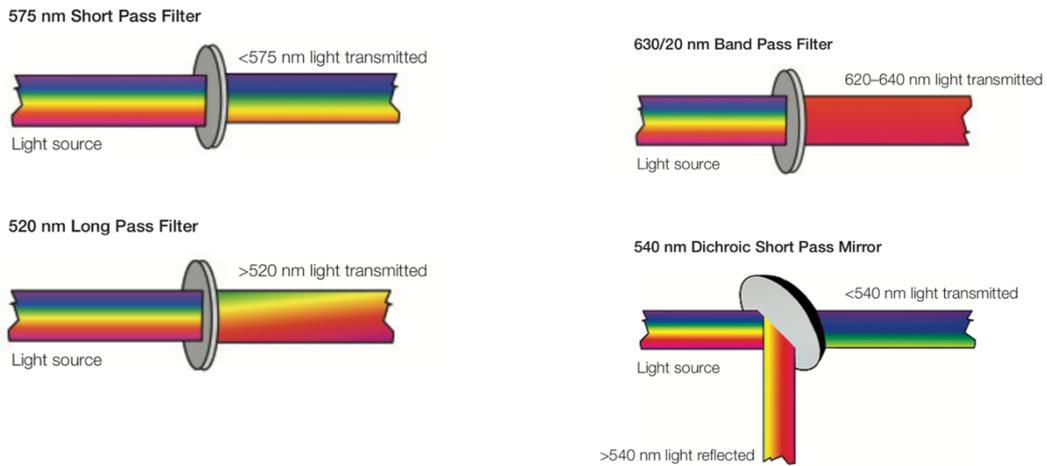

Figure 7. Different types of optical filters [4].

The overview of a traditional flow cytometer setup is presented in The.

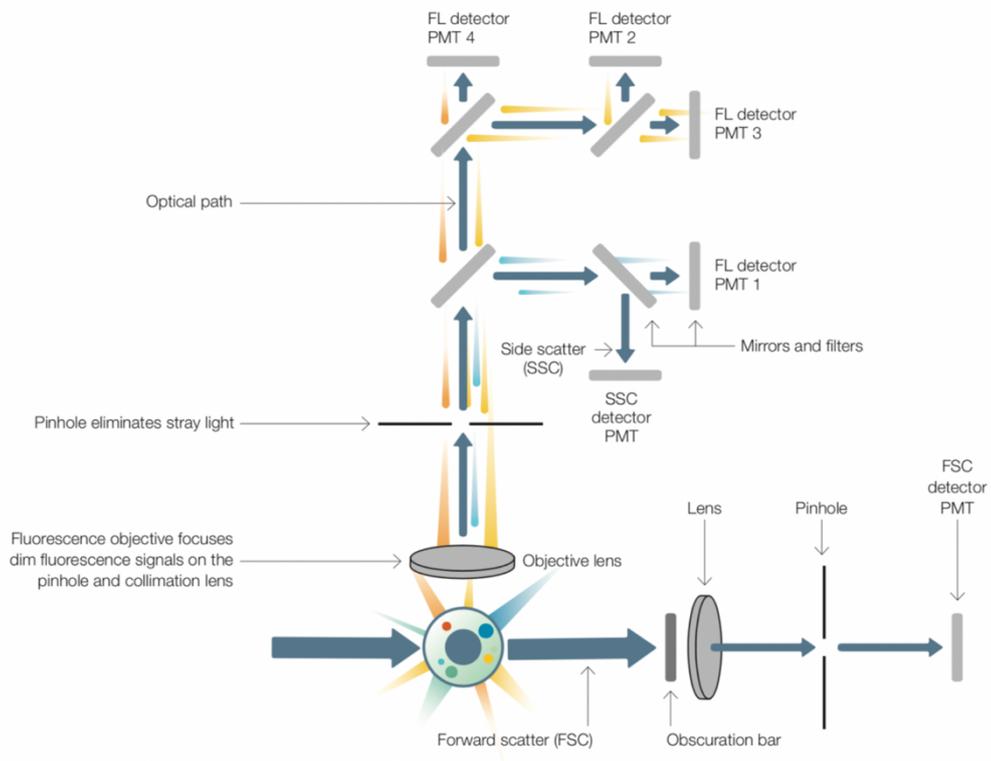

FL: fluorescence; PMT: photomultiplier tube; SSC: side scatter: FSC, forward scatter: blue arrow, light path

Figure 8. Typical flow cytometer setup[4].

## 2.1.3.3 Electronics

The light must be converted to electronic signals for the computer to analyse. This process happens in the electronic system, where the light signals of the various channels are combined and analysed, and the physicochemical characteristics of the cells in the sample population are obtained.

Filters and mirrors separate optical signals according to different wavelengths; then, the optical signals enter their respective channels, specifically their respective photomultiplier tubes (PMTs). The PMT has two main functions. First, it converts the optical signal into an electronic signal. Flow cytometry relies on computers to process and analyse data, and the computer requires an electronic signal to process the information. The PMT also amplifies, according to a certain ratio, the signal when converting an optical signal into an electronic signal. Flow cytometry analyses and processes data in units of single cells; however, the scattered light and fluorescent signal of one cell are weak. Moreover, the flow cytometer collects only one direction when collecting light signals instead of collecting all the scattered light and the fluorescent signal. Thus, if the signal is not amplified, the computer may not be able to effectively analyse it. The PMT be is connected to the optical system and the computer system, playing a key role as a bridge between the two.

The PMT is actually a channel. The channels of the flow cytometer can be divided into a scattered light channel and a fluorescent channel, depending on the nature of the light signal. The FSC channel and the SSC channel are each assigned to a photomultiplier tube. Basically, all flow cytometers have these two channels because they describe two very basic and important kinds of physical information about the cell.

A PMT converts the scattered light signal and the fluorescent signal into electronic signals which the computer system receives in the form of electronic pulses or waves and then analyses. As shown in Figure 9, there are three main ways to compare the electronic waves: the height (H), the width (W), and the area (A) of the electron wave. Stronger optical signals, after being converted into an electronic signal by the same multiple, are greater in height and width. However, the area represents the magnitude of the electron wave and is more accurate than the width and height.

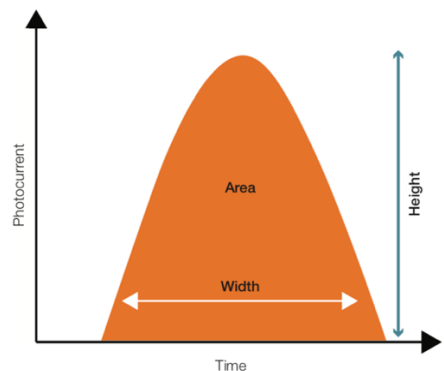
Figure 9. The meaning of A, H, W in Channels [4]

## 2.1.4 Application of Flow Cytometry

Advances in flow cytometers has expand the applications of flow cytometry. Flow cytometry is used in the fields of immunofluorescence, cell cycle kinetics, cell kinetics, parasitology and bioterrorism. One important application of flow cytometry is to detect specific cell populations by their phenotypic properties, such as immune cells in the blood to diagnose blood-related diseases [5].

## 2.1.5 Special Flow Cytometry

### 2.1.5.1 Mass cytometry (CyTOF)

In traditional flow cytometry, antibodies are labelled with fluorochromes; however, in mass cytometry, cellular targets are labelled with metal-tagged antibodies [6][7]. The largest advantage of mass cytometry is that it does not require the use of compensation to solve the problem of overlapping fluorescence emission spectra. Thus, mass cytometry can more easily detect rare cell populations than conventional flow cytometry. However, the data acquisition rate in mass cytometry is lower (approximately 500 cells per second) than in conventional flow cytometry and requires more stringent experimental conditions.

### 2.1.5.2 Imaging Flow Cytometry (IFC)

Imaging flow cytometry (IFC) combines the advantages of traditional flow cytometry while also capturing real images of the cells [8]. It is more powerful than traditional flow cytometry and provides information on the morphological interpretation of cells. However, IFC still has many challenges to overcome, such as its slow data acquisition rate (5000 evts/sec) and improvements to its image quality.

## 2.1.6 Software tools

There are more than 50 open and free software tools for flow cytometry, this section focuses on three software tools or platforms used for data collection and automated gating algorithm implementation: FlowRepository, R/BioConductor and GenePattern. FlowRepository is mainly used for data storage and download; R/BioConductor provides useful packages to implement automated-gating algorithms; and GenePattern is an online platform that provides many modules for flow cytometry analysis but without programming skills.

### 2.1.6.1 FlowRepository

FlowRepository [9] is an online database where users can download MIFlowCyt-standard based flow cytometry data. Experimental conclusions, as well as purposed and used data in published peer-reviewed journals can be accessed easily in FlowRepository. FlowRepository can be served as an extended Cytobank, but FlowRepository has specific features unavailable in Cytobank, and vice versa. Users can browse public datasets, community datasets and most popular datasets easily, as shown in Figure 10.

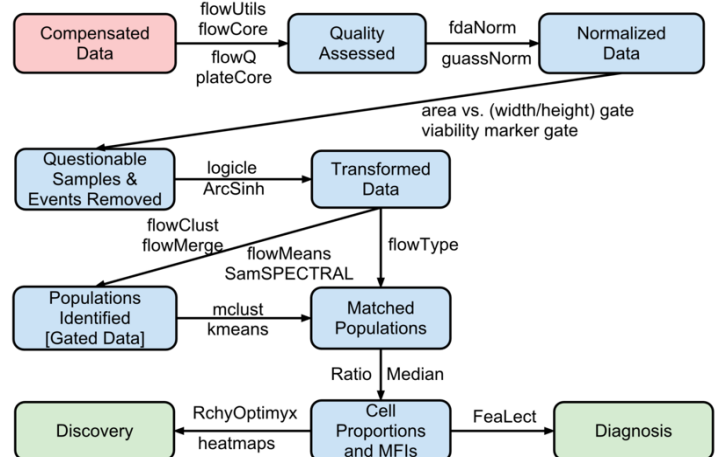

Figure 10. FlowRepository [9]

### 2.1.6.2 R/BioConductor

BioConductor [10] includes various automated-gating algorithms that require few R programming skills. To use most automated-gating algorithms, data should be transformed to text format (.csv or .xlsx) before being imported. Currently, more than 40 R/BioConductor have been used for flow cytometry data analysis. One of the advantages of R/BioConductor is that it easier to reproduce and it decomposes the problem into small modules. Figure 11 shows an example pipeline of flow cytoemtry analysis and corresponding required packages.

Figure 11. An example pipeline for analysis FCM data[11].

### 2.1.6.3 GenePattern

One limitation of R/BioConductor is it needs programming skills. GenePattern makes up the shortcoming R/BioConductor that provides more than 200 convenience analytical tools of gene expression, proteomics and other data [12]. Its web-oriented interface offers simple access to these tools and permits the formation of automated analysis pipelines that facilitate reproducible research. Additionally, GenePattern Flow Cytometry Suite [13] was created in a bid to bring advanced flow cytometry tools of data analysis to experimentalists that do not have programmatic skills. It comprises more than 30 open-source modules that cover methods from primary processing of FCS files to advanced algorithms for the automated identification of normalisation, cell populations and quality assessment. The pipeline of GenePattern Flow Cytometry Suite as shown in Figure 12.

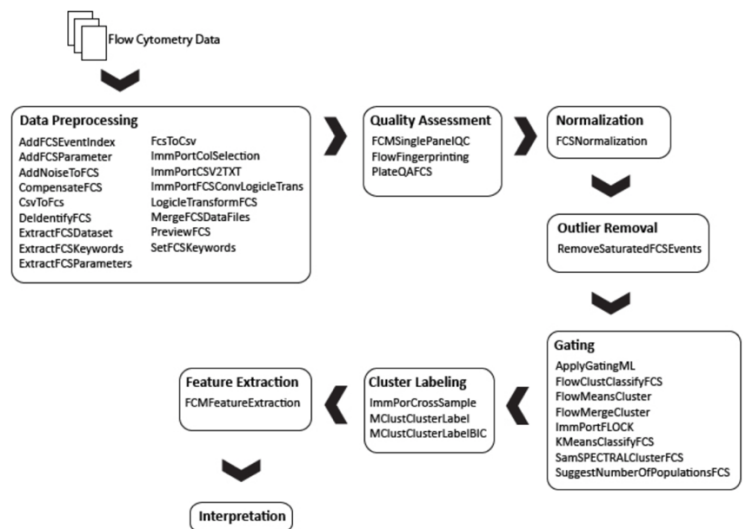

Figure 12. GenePattern Flow Cytometry Suite pipeline [13]

In addition to some of the software tools mentioned above, other software tools or programming languages can be used in flow cytometry data analysis, such as Java, Python, MATLAB and C++. However, such tools require higher programming experience and would not be suitable for every researcher.

## 2.2 Data Descriptions

### 2.2.1 Data Types

A better understanding of data types is a crucial prerequisite for building machine learning models. The data can be roughly divided into categorical data, numerical data, multimedia data (e.g., image, audio, video and text) and time series data. This section only covers a portion of these data types.

Categorical data is obtained by classifying or grouping according to certain attributes of a phenomenon. There are no quantitative relationships or differences between categorical data. Unlike categorical data, numerical data can be measured and counted, and it can exhibit quantitative relationships and differences.

Machine learning algorithms usually only accept numerical data. Thus, categorical data often needs to be transformed into numerical data during the data pre-processing step.

Time series data is collected at different time points to describe phenomena as a function of time. This type of data reflects the state or extent of how something changes over time. The study of time series is generally used in econometrics and is important in the field of finance.

Text data refers to any character that cannot participate in arithmetic operations. Before it can be used to train machine learning models, text data must be transformed to vectors. Image data refers to a set of grey values of pixels represented by numerical values. Audio data is collected by performing analogue-to-digital conversion (ADC) on the continuous analogue audio signal from a device such as a microphone to obtain audio data at a certain frequency. Digital-to-analogue conversion (DAC) uses the playback of digital sound to convert audio data into analogue audio signal output. Figure 13 depicts the aforementioned data types.

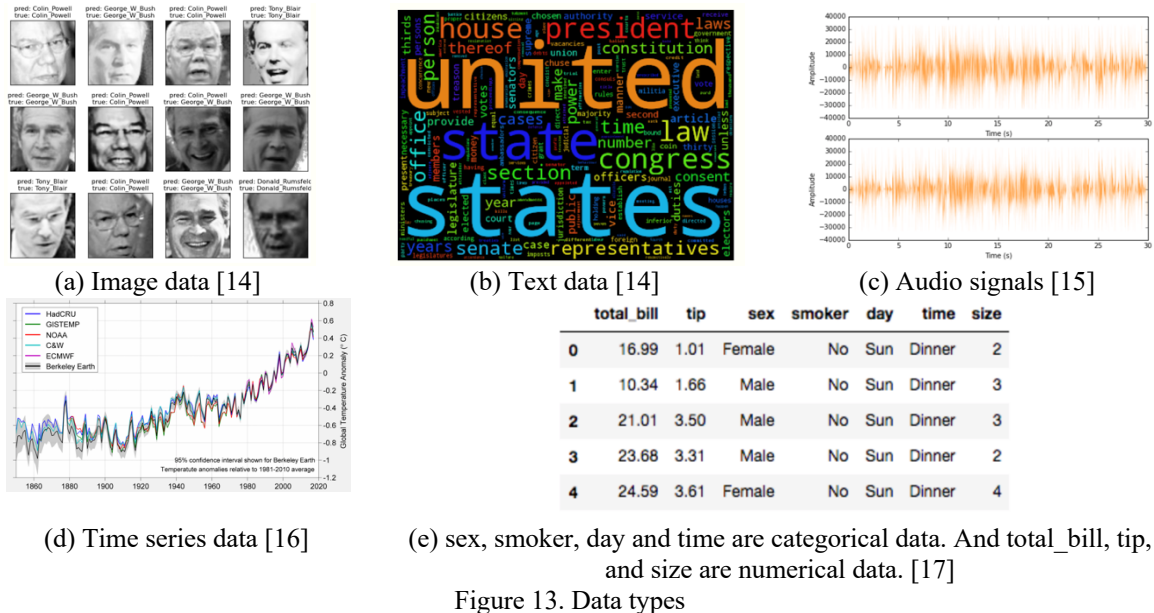

(a) Image data [14]    (b) Text data [14]    (c) Audio signals [15]

(d) Time series data [16]

(e) sex, smoker, day and time are categorical data. And total_bill, tip, and size are numerical data. [17]

Figure 13. Data types

The data generated from traditional flow cytometry is numeric data, and imaging flow cytometry generates image data. Understanding these flow cytometry data types can be a direction for designing machine learning models.

### 2.2.2 Flow Cytometry Data

#### 2.2.2.1 Characteristics

The size of flow cytometry datasets is increasing rapidly due to the improvement of flow cytometers. Since 2015, there are now more than $4^{11}$ flow cytometry datasets, which include more than 77,000 samples and 1,000,000 cells, as shown in Table 1.

Table 1. Fast Growing of Flow Cytometry Data [18][19][20]

|  | 1985 | 2012 | 2015+ |
|---|---|---|---|
| **Samples** | 1 | 466 | 77,000 |
| **Dimensions** | 5 | 13 | 40+ |
| **Cells** | 50,000 | 400,000 | 1,000,000 |
| **Datasets** | $2.5^5$ | $2.5^9$ | $4^{11}$ |

Understanding the features of flow cytometry data aids in the design of appropriate automated rating algorithms. Flow cytometry data is multi-dimensional, unbalanced and of different sizes; each data file has anywhere from hundreds to millions of events and could also contain many noises, outliers, dirt and dead cells. Cell populations are divergent in density and shape. While some cell populations are considerably sparse, even in a two-dimensional space composition, others could be very close or far apart in terms of the size. Events that accumulate on axis could alter data distribution positions. Therefore, the features of flow cytometry data could cause difficulties when utilizing traditional flow cytometry analysis methods.

### 2.2.2.2 Data File Formats

The lack of standardisation curbs collaborative chances of recreating experimental methods and outcomes. Thus, several FCM data standards have been developed to overcome this issue, such as Flow Cytometry Standard (FCS), MIFlowCyt Standard, ISAC Classification Results File Format (CLR) and Gating-ML File Format.

#### 2.2.2.2.1 Flow Cytometry Standard (FCS)

The first version of FCS was published by the International Society for Advancement of Cytometry (ISAC) in 1984 [21][22]. To adapt it to take into account the advances in flow cytometers, update versions of FCS have been developed, such as FCS 2.0 [23], FCS 3.0 [24], and the latest version, FCS 3.1 [25].

The basic structure of an FCS file includes three segments: HEADER, TEXT, and DATA. The specific version of FCS used can be found in the HEADER segment. The TEXT segment includes several keyword-value pairs that describe the DATA segment format and encoding. The DATA segment using a list mode format to store and describe the fluorescence and scatter channels for each event.

#### 2.2.2.2.2 Minimum Information about a Flow Cytometry Experiment (MIFlowCyt) Standard

One disadvantage of FCS is it does not fully utilize flow cytometry data [26]. Thus, additional FCM data standards have been published to make up for FCS's shortcomings. Minimum information about a Flow Cytometry Experiment (MIFlowCyt) standard improves the understanding, reuse and unambiguous interpretations of flow cytometry data by outlining the minimum information required to record and report a flow cytometry experiment [27]. FlowRepository [9] requires users to use the MIFlowCyt standard to annotate their dataset [28], and many journals also require the MIFlowCyt standard, such as *Nature* and *PLOS*.

#### 2.2.2.2.3 Gating-ML File Format

FCS uses the $GATING keyword to encode the information of a one-dimensional range gate or two-dimensional polygon gate. However, FCS focuses on handling raw data, and the $GATING keyword is designed for the data acquisition process [29]. Thus, the gating descriptions used for the post-acquisition analytical process captured by FCS are not enough and cannot be shared for independent researchers and different software. This shortcoming will limit the full use of flow cytometry data [30]. To overcome this disadvantage, Gating-ML, an Extensible Markup Language (XML)-based mechanism, to encode unambiguous gating descriptions used for post-acquisition analysis was developed by ISAC in 2008 [29]. FCS and Gating-ML complement each other, and the latest version is Gating-ML 2.0 published in 2013 [30].

### 2.2.2.2.4  ISAC's Classification Results File Format (CLR)

Classification Results File Format (CLR) [31] can be seem as clusters results format based on CSV spreadsheet format. The increased in high-throughput and high-content flow cytometry data [32] has led to the development of various computational-based automated algorithms to supplement manual gating, the traditional methods of flow cytometry data analysis. However, Gating-ML cannot capture obtain the results of the automated methods due to the results often have no unambiguous boundaries [31]. Thus, ISAC develops another file format CLR to overcome this shortcoming. The basic structure of CLR format FCM data as shown in Figure 14, the rows correspond to event, and columns to class. CLR use probability to represent as event belongs to a particular class.

|         | Class 1 | Class 2 | | Class n |
|---------|---------|---------|---|---------|
| Event 1 | 1       | 0       | | 0       |
| Event 2 | 0.2     | 0.6     | | 0.1     |
|         |         |         | |         |
| Event n | 0.3     | 0.7     | | 0       |

Figure 14. An example of CLR format FCM data

## 2.3  Data Analysis in Flow Cytometry

## 2.3.1 Framework

Flow cytometry data analysis typically has five steps [33]:
(1) Data pre-processing: Outliers (e.g., dead cells) are removed from the raw data, as are overlapping fluorescence signals and transformations.
(2) Automated gating: Homogeneous cell populations are identified by supervised or unsupervised learning.
(3) Cluster labelling: This is an optional step and can be done during automated gating. The goal is to find homogeneous cell populations between samples.
(4) Feature extraction: Some features can be extracted from gated data, such as the number of cells per cluster or the parameter used for gated data.
(5) Interpretation: The results of automated gating are interpreted. The interpretation results depend on the goal of the research.

The basic flow cytometry data analysis framework is shown in Figure 15.

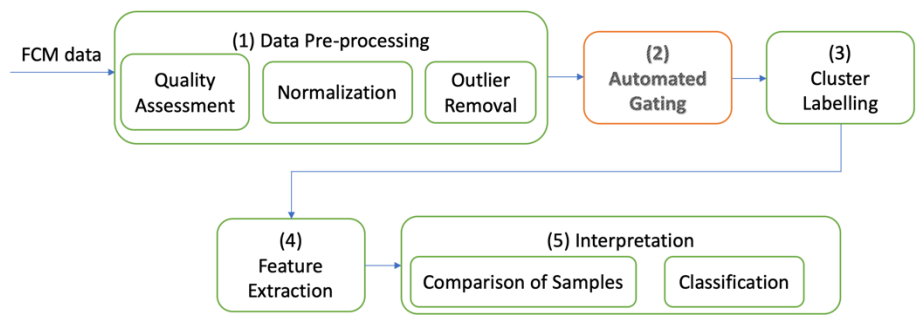

Figure 15. An example of Flow cytometry data analysis framework [33]

Most flow cytometry data research has focused on the development of automated gating techniques, especially unsupervised automated gating techniques, as shown in Figure 16.

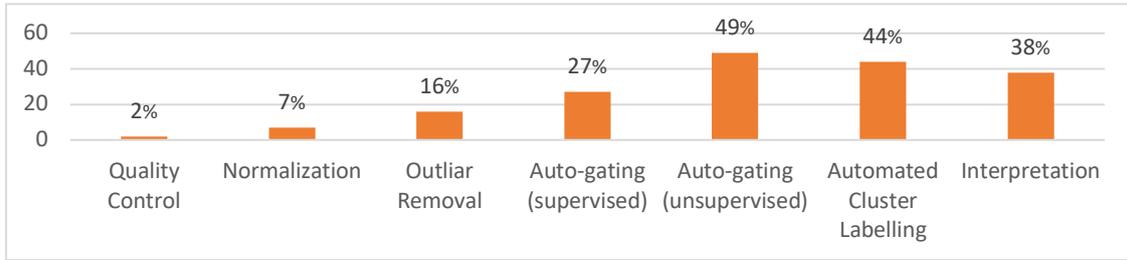

Figure 16. Percentages of current studies in different data analysis components [33].

## 2.3.2 Data Pre-processing

Flow cytometry data need to be pre-processed before analysis to obtain accurate and reliable analysis results. Traditional flow cytometry data pre-processing includes four steps: compensation, transformation, quality control and normalisation.

### 2.3.2.1 Compensation

Spectral overlap often occurs when multicolour fluorescence markers are used during analysis. The overlap between wavelengths of two fluorochromes is called spillover. Spillover causes the display of incorrect or non-existent fluorescence signals [34]. The method used to avoid spillover is termed compensation [35]. Compensation can be simply thought of as a process used to manually or automatically choose fluorochromes that do not overlap. Manual compensation often causes overcompensate problems; automatic compensation can effectively avoid such problems [34]. Compensation can be accomplished by professional flow cytometry software, such as flowCore [36]. Figure 17 shows a comparison between compensated data and data without compensation.

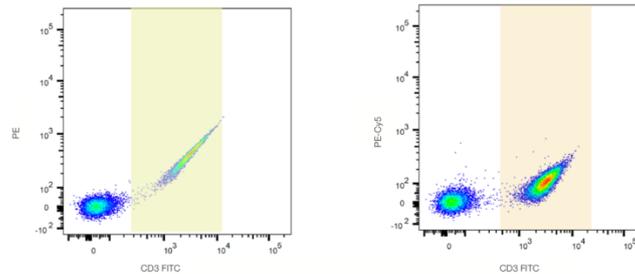

Figure 17. Comparison of compensated data (right) and data without compensation (left).

### 2.3.2.2 Transformation

Transformation is a process to transform the data into a suitable scaling. The original data are usually expressed as log-normal [37] and need to be transformed to a different scale, such as a logarithmic scale, Logicle [38][39], Hyperlog [40] or Box-Cox [41]. Further details regarding transformation can be found in Section 2.3.4.

### 2.3.2.3 Quality Control

During data pre-processing, outlier events (e.g., dead cells) need to be removed from each sample. In addition, the outlier samples also need to be removed from the dataset. This process is called quality control. Commonly quality control methods include the Kolmogorov–Smirnov test [42], Grubbs's test and probability binning [43].

### 2.3.2.4 Normalization

Non-biological variances (e.g., experimental variances) can influence the results of studies. Normalisation is a method of standardising the data by reducing the variance [44], which is a crucial step for flow cytometry data analysis. The basic normalisation types include standard score and min–max scaling; the equations are shown as follows:

$$\text{Standard score} = \frac{X - \mu}{\sigma}$$

$$\text{Min-Max scaling} = \frac{X - X_{min}}{X_{max} - X_{min}}$$

### 2.3.3 Data Presentation

A flow cytometer can analyse and sort cells at high speed, with a modern flow cytometer being able to analyse and sort tens of thousands of cells per second. Each analysed cell can obtain several aspects of information. The most basic one is the value of FSC and SSC. If cells are labelled with fluorescein-conjugated antibodies at the same time, numerous fluorescent channel information will be obtained. For example, when the sample cells are labelled with four fluorescein-conjugated antibodies, each cell contains six channels' data. If 10,000 cells are analysed in one second, then 60,000 pieces of information need to be stored and analysed in one second. The output data from flow cytometry can be described as a matrix or array, where the columns are the individual events and the rows are the parameters of flow cytometry, namely FS, SS, and/or fluorescence, as shown in Figure 18.

Figure 18. An example of List-Mode FCM data, which rows represented by cells (events), and scatter and fluorescence channels forming the columns [11].

This type of data is easy to process, but it lacks intuitiveness. Thus, many ways can be used to display such a large amount of information fully and intuitively, with the most commonly used being histograms, scatter plots, and contour maps. Figure 19 presents the eight ways to display flow cytometry data.

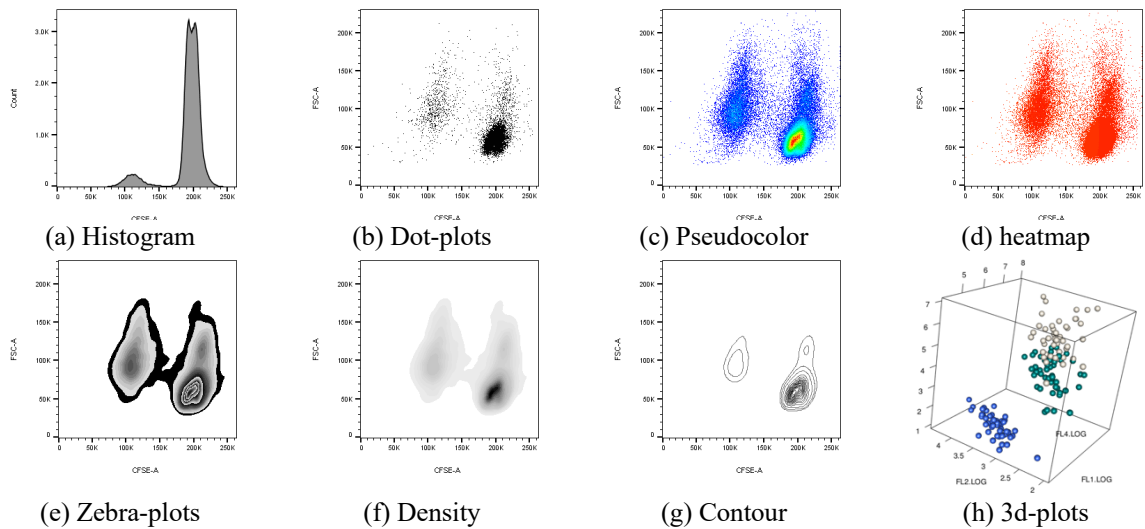

| (a) Histogram | (b) Dot-plots | (c) Pseudocolor | (d) heatmap |
| (e) Zebra-plots | (f) Density | (g) Contour | (h) 3d-plots |

Figure 19. Eight commonly ways to present flow cytometry data

### 2.3.3.1 Channels

Channels of flow cytometers can be divided into scattered light channels and fluorescent channels, and the scattered light channels include FSC channels and SSC channels, which are contained in most the flow cytometers. The values of FSC and SSC represent the size and granularity of the cell, the two basic physical characteristics of the cells. Thus, the samples' FSC-SSC scatter plot is also known as the "physical map" of the sample cells.

The fluorescent channel represents not the unique physical characteristics of the cell but its chemical characteristics. Should it be necessary to detect whether a cell expresses a certain CD molecule in the sample, it is required to label the antibody of the CD molecule conjugated with fluorescein. Then, the cell expressing the CD molecule binds to the fluorescein-conjugated antibody. The corresponding laser excites the fluorescein to produce fluorescence, and the corresponding fluorescent channel receives the fluorescent signal through the optics system. The stronger the fluorescent signal, the more fluorescein is bound on the cell, and the more the CD molecule is expressed on the cell surface. Thus, the relative amount of the CD molecule can be judged based on the intensity of the fluorescent signal.

### 2.3.3.2 Histogram

The relationship of a channel's value and cell counts can be presented by a histogram. Thus, the histogram displays the distributions of the event where the x-axis represents the channels' value and the y-axis represents the cell counts. A cell population of the same properties appears in the form of a normal distribution curve in the histogram. Histograms are useful when same antigen or marker is analyzed through multiple samples [45], and comparing the total number of cells in multiple samples that have been stained with the same marker or antigen of interest [46].

### 2.3.3.3 Dot-plots

A histogram can only represent information for one channel, while a scatter plot can represent information for two channels simultaneously. Thus, scatter plots can be more intuitive in representing cell information and are more commonly used when compared with the histogram. In the scatter plot, the cell populations that have a shape of a circle with a typical radial distribution have the same properties; those that have a long or elliptical shape are likely to have a strong difference in expression. However, it is not as accurate to judge whether a certain characteristic of a cell population is different based on a scatter plot. There are many factors that influence the shape of the cell population in the scatter plot. Specifically, the difference between the strong and the weak expression as well as the fluorescence compensation between the channels will directly affect the shape of the cell population. Thereby, the original circular shape of the cell population will become an elongated shape.

### 2.3.3.4 Contour map

The contour map is similar to the scatter plot, as it can also display two channels' information simultaneously; however, it is in the form of a geographic contour map. The geographical contour map uses a closed loop to represent the same altitude. The more the loops gather, the faster the altitude changes. The central area of the loop indicates the highest or lowest altitude. With cells, the contour map shows the density of cells by means of the same features as a geographical contour map. The loop of the contour map represents the region with the same cell density. Therefore, the more the loop gathers, the more the cell density changes in this region. The most sparsely populated area is represented by scatter, and the central region of the loop represents the centre of cell aggregation.

The contour map is not as intuitive as the scatter plot, and it is relatively insensitive to cell numbers, which could be misleading for some, as some cells, in particular rare cells, are excluded in the plots [46]. However, a contour map can more intuitively reflect the concentration point of the cell population with the central region of the equal density loop representing a concentration point of a cell group, generally representing a cell population. Thus, in this case, the contour map is more intuitive than the flow scatter plot to reflect the cell population.

Other than using a graph to present several parameters of single cell population, multigraph is another useful option to display all parameters regarding cell population. Multigraph overlays or N × N plots (Figure 20) allow a visual comparison of the same parameters of multiple cell subpopulations of the same sample or of the same parameters of one subpopulation from different samples [46].

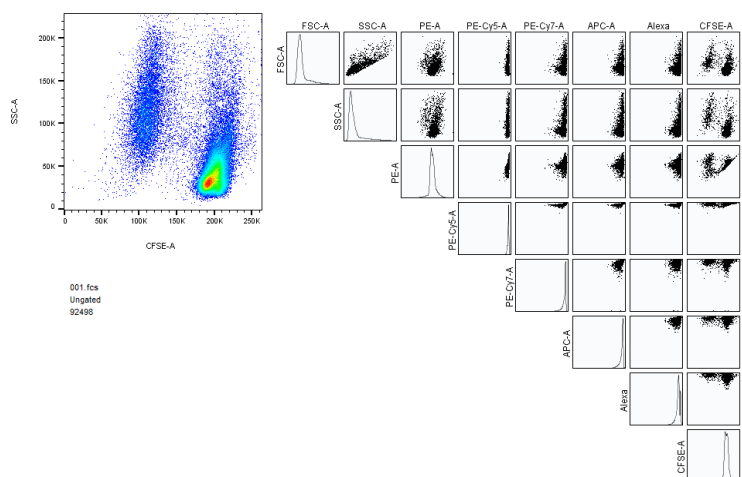

Figure 20. N × N plots display all parameters of selected populations.

## 2.3.4 Data Scaling

The FSC value and SSC values of different cell populations can differ by up to several times. However, the fluorescence signal strength is generally very different, sometimes differing by up to a few hundred times. Therefore, the FSC and SSC signals are often scaled by linear scaling, while fluorescence signals are scaled by logarithmic scaling. When the fluorescence signal values are similar, they can also be scaled by linear scaling [47]. In linear scaling, the FSC, SSC or fluorescence channel number can be converted to a linear value by the equation $channel\ number = \log(linear\ value) \times 256$ or by using the equation $10^{\frac{channel\ number}{256}} = linear\ value$ to convert a linear value to a channel number [46]. However, compensated data usually have low means and normal distributions [38]. Logarithmic scaling is often unable to correctly represent unstained cell populations, low fluorescence signals or normally distributed cell types because it compresses the channels of visual space as the scale increases [38][48]. To overcome this drawback, a new scaling method called biexponential scaling has been developed [49], as shown in Figure 21.

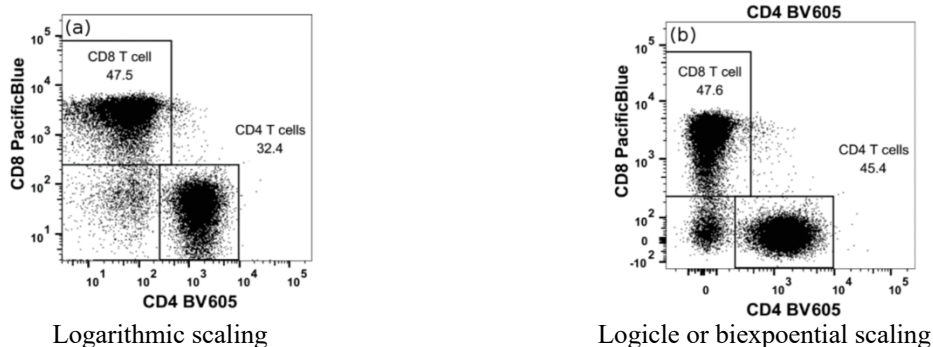

Logarithmic scaling       Logicle or biexpoential scaling

Figure 21. Data Scaling

## 2.3.5 Cell Population Identification

### 2.3.5.1 Manual Gating

Gating is a process used to identify interesting cell subpopulations by the use of "gates". This process can be conducted by professional software, such as FCS Express [51], FlowJo [50] and flowCore [36]. Clinical decisions are made based on the shape of gated data, which can be in the form of one- or two-dimensional plots. Figure 22 shows two representations of gating; the interesting cell subpopulations can either be surrounded by a boundary or highlighted using the same colour.

However, conventional flow cytometry data analysis is highly dependent on manual gating and is mainly suitable for low-dimensional data with a limited number of cells. A simple example of manual gating is shown in Figure 23. The results of manual gating are often quite subjective due to human intervention—it is difficult to segregate highly overlapping cell populations by only looking at the data plots. Therefore, it is necessary to develop suitable computation methods to overcome the drawbacks of manual gating.

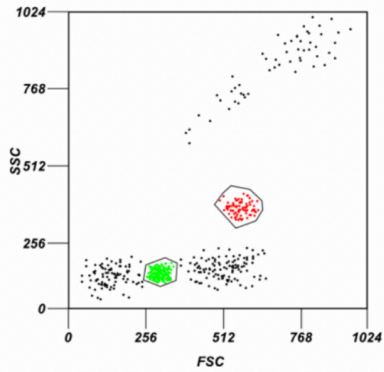 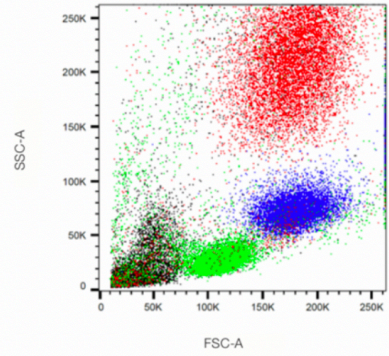

(a) A mock example of gating. In the image two subpopulations of cells are gated based on their forward and side-scatter intensity profiles.

(b) The same population of cells are highlighted by a cohesive colour scheme when analysing two different channels measuring fluorescent intensity in the image above.

Figure 22. Examples of gating [52]

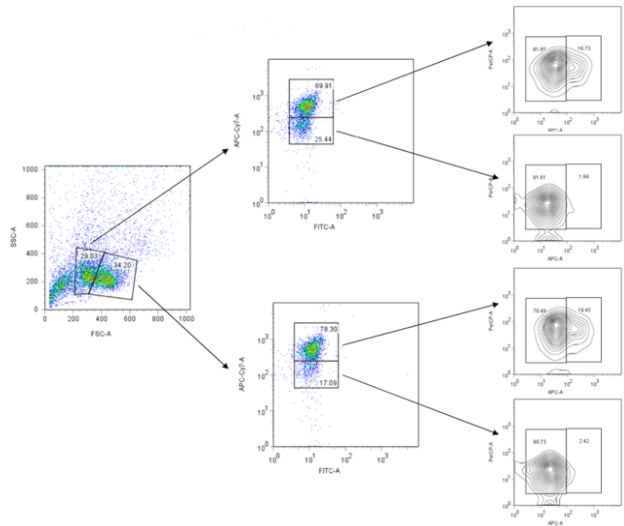

Figure 23. Traditional flow cytometry analysis

### 2.3.5.2 Automated Gating

Most current automated gating algorithms are based on machine learning, which can be roughly divided into three groups: supervised learning, unsupervised learning and deep learning. As Figure 16 shows, unsupervised automated gating occupies the majority of existing automated algorithms. Further details regarding automated gating algorithms based on machine learning are introduced in Chapter 3.

Table 2 lists a summary of the flow cytometry data analysis steps and required software.

Table 2. Steps of the FCM data analysis framework and associated software or packages [53]

| Stage | Step | Software / R packages |
|---|---|---|
| Pre-processing | Compensation | flowCore, flowUtils |
| | Transformation | flowCore, flowTrans |
| | Normalization | flowStats (fdaNorm, gaussNorm) |
| | Quality Assessment | QUALIFIER, flowAI, flowClean |
| Gating | Sequential manual gating | FlowJo, flowCore |
|  | Automated gating | Supervised algorithms: flowDensity, DeepCyTOF, flowLearn, OpenCyto |
|  |  | Unsupervised algorithms: flowMeans, SPADE, Citrus, FlowSOM, cytometree, ECLIPSE |
| **Cluster Labelling** | Population matching | flowMatch, flowmap-RF |
| **Interpretation** | Diagnosis and discovery | flowType, MetaCyto, CytoCompare |
| **Extra processing steps** | Visualization | flowViz, ggCyto, RchyOptimyx, SPADE, Citrus, t-sne |

# 3 Review of Machine Learning Technique

The information of raw flow cytometry data can be gained using machine learning techniques [54]. Fundamentally, machine learning is optimisation, as it optimises loss functions. There are four basic types of machine learning: supervised learning, unsupervised learning, reinforcement learning and deep learning.

## 3.1 Supervised Learning

In order to design a suitable machine learning algorithm, it is necessary to know not only which type of data is involved but also the characteristics of the different models. Supervised learning models can mainly be divided into two groups: regression and classification. Regression models predict unknown labels based on the past and inferences. Classification models use parts of labelled data as training data and divides unlabelled data into classes belonging to training data [44]. The difference between classification and regression also can be seen in Figure 24.

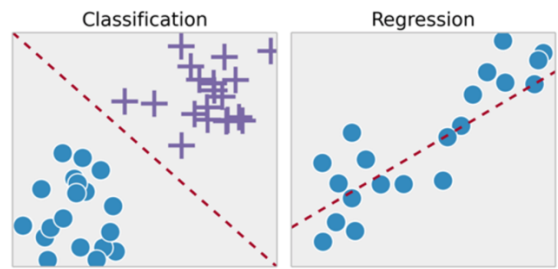

Figure 24. The difference between classification and regression. Classification separates the data, Regression fits the data [55]

When using supervised learning algorithms, the labelled data often needs to be divided into three parts: training data, testing data and validation data. Models use training data to learn the relationship between labels and observations. Testing data is used to evaluate the performance of the model, and validation data is used to adjust the parameters used in a model. Therefore, if the flow cytometry dataset is tagged with cell types per event (which cluster the event belongs to), then supervised learning can aid in the diagnosis or the discovery of new cell types.

However, supervised learning is unsuitable for flow cytometry data analysis because it requires large amounts of data with known and correct labels for training stages. Current fully gated flow cytometry datasets, such as DLBCL and GvHD, are all gated by humans, which means the labels may not all be correct. Therefore, a possible method to reduce the limitations of supervised learning for flow cytometry data analysis would be by using supervised learning to simulate the manual gating strategy rather than using the gating results provided by humans.

FlowDensity uses the density distribution of markers to estimate the gating region and stimulate the manual gating strategy by a sequential bivariate gating algorithm [56]. However, flowDensity requires the gating strategies are known before being implemented.

## 3.2 Unsupervised Learning

The difference between supervised and unsupervised learning is the latter do not require the data must be labelled. Therefore, the internal properties of data are used for unsupervised learning. Most automated gating algorithms are based on unsupervised learning, and the typical used unsupervised learning algorithms for flow cytometry analysis is clustering [33]. Clustering is similar to classification, except it do not need data must be labelled. In this section, the commonly types of clustering will be discussed: hierarchical, density-based, partitioning, grid-based and model-based clustering.

### 3.2.1 Hierarchical Clustering

Hierarchical clustering algorithms cluster data based on its hierarchical relationship [57]. Hierarchical clustering can be grouped into two categories based on its order of merge similar clusters: agglomerative and divisive hierarchical clustering [58]. Agglomerative hierarchical clustering merges similar clusters from bottom to up, which think each data as an individual cluster in the first level, then merge most similar data as a group in the second level, and so on, till the data cannot be merged anymore. Divisive hierarchical clustering performs the reverse process. The grouping steps of both agglomerative and divisive clustering are presented in Figure 25. Agglomerative and Divisive clustering. Generally, agglomerative clustering can group small clusters well, while division clustering is good at group large clusters.

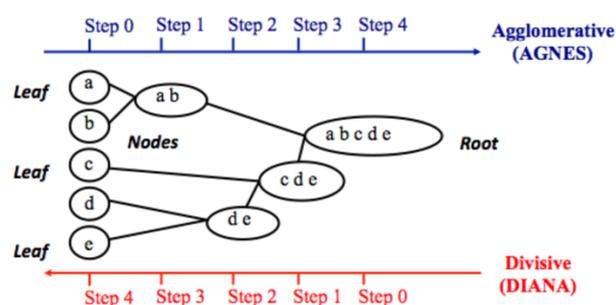

Figure 25. Agglomerative and Divisive clustering [59]

The hierarchical clustering algorithms also can be grouped in terms of the linkage (a similar measurement), as presented in Figure 26.
- Average linkage: the average distance are used to estimate the distance between two clusters;
- Complete linkage: the longest distance are used to measure the distance between two clusters;

- Single linkage: the shortest distance are applied for measure the distance between two clusters.

The commonly used hierarchical clustering algorithms include BIRCH [60], ROCK [61] and CURE [62]. In flow cytometry data analysis, a commercial software ADICyt [63] clusters data based on hierarchical clustering, and merges data based on entropy.

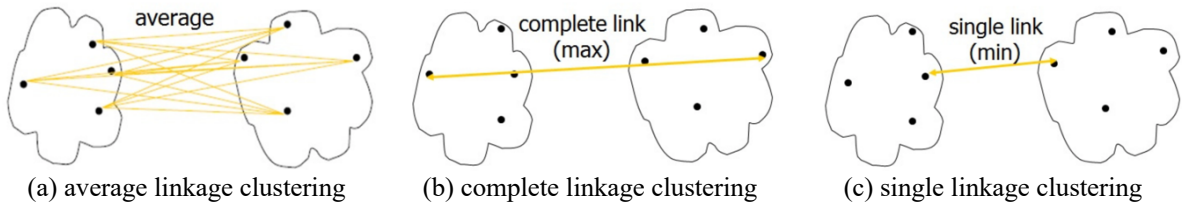

(a) average linkage clustering  (b) complete linkage clustering  (c) single linkage clustering
Figure 26. Grouped hierarchical clustering based on linkage (Euclidean distance in this case).[64]

### 3.2.2 Partitioning Clustering

Partitioning clustering is also called flat clustering, it is simply partitioning the data into $k$ disjoint groups based on the corresponding center of data points [65]. Partitioning clustering often faster than hierarchical clustering, but more ineffective than hierarchical clustering [65]. The popular partitioning clustering algorithms include K-medoids [66], K-means [67], CLARANS [68], PAM [69] and CLARA [70]. In flow cytometry data analysis, flowMeans [71] modified K-means to identify cell populations. Figure 27 depicts an example of partitioning clustering.

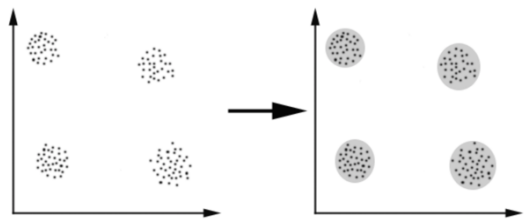

Figure 27. An example of partitional clustering [72]

### 3.2.3 Density-based Clustering

The idea of density-based clustering is the density of clusters are higher than outliers [73]. Most partitional clustering cannot group concave data, but density-based clustering can discover the arbitrary shape clusters and robust to noise, as Figure 28 shows. The typical density-based clustering algorithms include DBSCAN [74], OPTICS [75] and Mean-shift [76]. In flow cytometry data analysis, the density-based automated clustering algorithms include NMFcurvHDR [77], SPADE [78] and MMPCA [79].

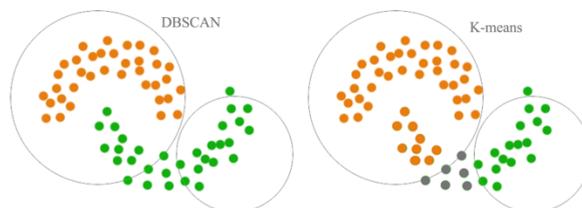

Figure 28. Grouping concave data based on density (left) and based on distance (right) [80]

### 3.2.4 Grid-based Clustering

The grid-based clustering quantizes the data element space into a finite number of cells to form a grid structure [81]. All clustering operations are performed on this grid structure [81]. Grid-based algorithms are often used in conjunction with density-based algorithms, such as CLIQUE [82]. A significant advantage of this type of algorithm is that it runs very fast. Figure 29 presents an example of grid-based clustering. In flow cytometry, FLOCK [83] uses the idea of grid-based to partition and merge the data.

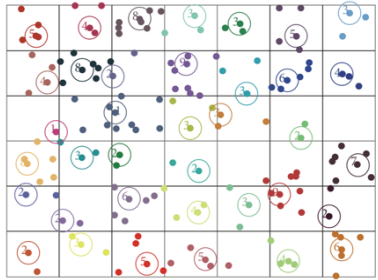

Figure 29. An example of grid-based clustering [84]

### 3.2.5 Model-based Clustering

The model-based clustering attempts to assume a mathematical model for the dataset, such as statistical models and neural network models, and then find the best fittings [81]. The typical model-based clustering algorithms based on statistical model include GMM [85] and COBWEB [86]. Model-based clustering is often used in flow cytometry data analysis: CDP [87] uses Bayesian nonparametric models; the parametric and multivariate mixture modelling is used in FLAME [88]; flowClust[41] and flowMerge [89] both applied t-multivariate modelling.

The pros and cons of previous mentioned clustering algorithms are summarized in Table 3.

Table 3. The summary of pros and cons for different types of clustering [90]

| Clustering Types | Pros | Cons |
|---|---|---|
| Hierarchical | Suitable for the dataset with arbitrary shape and attribute of arbitrary type, the hierarchical relationship among clusters easily detected, and relatively high scalability in general | Relatively high in time complexity in general, the number of clusters needed to be pre-set |
| Partitioning | Relatively low time complexity and high computing efficiency in general | Not suitable for non-convex data, relatively sensitive to the outliers, easily drawn into local optimal, the number of clusters needed to be pre-set, and the clustering result sensitive to the number of clusters |
| Density-based | Clustering in high efficiency and suitable for data with arbitrary shape | Resulting in a clustering result with low quality when the density of data space isn't even, a memory with big size needed when the data volume is big, and the clustering result highly sensitive to the parameters |
| Grid-based | Low time complexity, high scalability and suitable for parallel processing and increment updating | The clustering result sensitive to the granularity (the mesh size), the high calculation efficiency at the cost of reducing the quality of clusters and reducing the clustering accuracy |
| Model-based | Diverse and well-developed models providing means to describe data adequately and each model having its own | relatively high time complexity in general, the premise not completely correct, and the |

| | special characters that may bring about some significant advantages in some specific areas | clustering result sensitive to the parameters of selected models |
|---|---|---|

A comparative table for some popular clustering algorithms as shown in Table 4.

Table 4. Comparative study of some clustering algorithms [90]

| Types of Clustering | Algorithm | Time Complexity | Scalability | Suitable for large scale data | Suitable for high dimensional data | Sensitive of noise |
|---|---|---|---|---|---|---|
| Partition | K-means | Low | Middle | Yes | No | High |
| | PAM | High | Low | No | No | Little |
| | CLARA | Middle | High | Yes | No | Little |
| | CLARANS | High | Middle | Yes | No | Little |
| Hierarchy | BIRCH | Low | High | Yes | No | Little |
| | CURE | Low | High | Yes | Yes | Little |
| | ROCK | High | Middle | No | Yes | Little |
| Density based | DBSCAN | Middle | Middle | Yes | No | Little |
| Grid based | CLIQUE | Low | High | No | Yes | Moderate |

## 3.3 Reinforcement Learning

Reinforcement leaning enables data to train by itself, also known as self-training. Reinforcement learning is not a specific algorithm but a general term for a class of algorithms. The essence of reinforcement learning is to solve the problems associated with decision making, specifically to make decisions automatically and continuously. It consists of four elements: agents, environmental status, actions and rewards. The goal of reinforcement learning is to get the greatest number of cumulative rewards.

Reinforcement learning is not mature enough at present, and the application scenarios are relatively limited. The biggest application scenario is in gaming, but it is still unsuitable in the field of flow cytometry data analysis. Some complex reinforcement learning algorithms have a general intelligence that is able to solve complex problems to a certain extent, reaching human level in Go and video games [91].

## 3.4 Deep Learning

Deep learning is a subset of machine learning, and the relationship between deep learning, machine learning and artificial intelligence is depicted in Figure 30. One difference between traditional machine learning and deep learning is that deep learning expands upon the capabilities of traditional machine learning. Traditional machine learning performs poorly when trying to determine whether a result is correct or not, but deep learning can do that. A deep learning model may be constructed by using a large number of layers, and the basic network architecture includes convolutional neural networks, recursive neural networks and unsupervised pretrained networks [54].

Deep learning excels in many areas, such as image processing, pattern recognition and so on [92]. Thus, deep learning algorithms can be used for analysing imaging flow cytometry data and to diagnose diseases [93][94]. In addition, deep learning algorithms can be used for mass cytometry data, such as DeepCyTOF [92]. However, most deep learning algorithms require a large amount of labelled data, similar to supervised learning. Deep learning, especially, requires huge multidimensional datasets with a large number of data (cells) and features (markers), which current flow cytometry cannot provide [53]. Thus, the research of deep learning in flow cytometry still holds great potential for discovery.

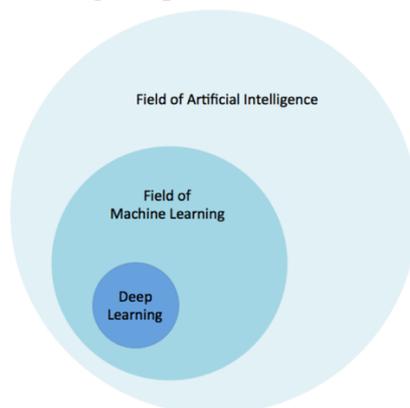

Figure 30. The relationship between Artificial Intelligence, Machine Learning and Deep Learning [29].

Table 5. Some current automated gating algorithms [102]

| Algorithm | Language | Brief Description | Ref |
|---|---|---|---|
| **Cell Population Identification** | | | |
| CDP | Python | Bayesian non-parametric mixture models, calculated using massivelyparallel computing on GPUs | [87] |
| FLAME | R | Multivariate finite mixtures of skew and heavy-tailed distributions | [88] |
| flowClust/Merge | R | t-mixture modeling and entropy-based merging | [41,89] |
| flowMeans | R | k-means clustering and merging using the Mahalanobis distance | [71] |
| NMF-curvHDR | R | Density-based clustering and non-negative matrix factorization | [77] |
| SamSPECTRAL | R | Efficient spectral clustering using density-based down-sampling | [95] |
| L2kmeans | JAVA | Discrepancy learning | [96] |
| Radial SVM | MATLAB | Supervised training of radial SVMs using example manual gates | [97] |
| SWIFT | MATLAB | Weighted iterative sampling and mixture modeling | [98] |
| FLOCK | C | Grid-based partitioning and merging | [83] |
| **Sample Classification** | | | |
| SWIFT | MATLAB | SWIFT clustering and support vector machines | [98] |
| flowCore-flowStats | R | Sequential gating and normalization and a Beta-Binomial model | [36] |
| flowPeakssvm Kmeanssvm | R | Kmeans and density-based clustering and support vector machines | [99] |
| flowType FeaLect | R | 1D gates extrapolated to multiple dimensions and bootstrapped LASSO classification | [100] |
| PBSC | C | Multi-dimensional clustering and cross sample population matching using a relative distance order | [101] |
| SPADE, BCB | MATLAB, Cytoscape, R/BioC | Density-based sampling, kmeans clustering, and minimum spanning trees | [78] |

# 4 Data Preparation and Evaluation Measurements

## 4.1 Real Datasets

The raw real data annotated with the MIFlowCyt standard could be accessed from FlowRepository [9] using the corresponding repository IDs: GvHD (FR-FCM-ZZY2), DLBCL (FRFCM-ZZYY), WNV (FR-FCM-ZZY3) and HSCT (FR-FCM-ZZY6).

## 4.1.1 Diffuse Large B-cell Lymphoma (DLBCL)

The dataset is provided by BCCRC, and contains 30 samples from patients with DLBCL (diffuse large B-cell lymphoma). The cells were stained for three antibodies (markers), CD3, CD5, and CD19. The samples in this dataset are containing both a small number of events (less than 2000) and a large amounts of events (more than 20000). The scatter plot of this data sets is depicting in Figure 31.

The details of DLBCL dataset can be seen in Appendices 2: Details of Real Datasets, Table 30.

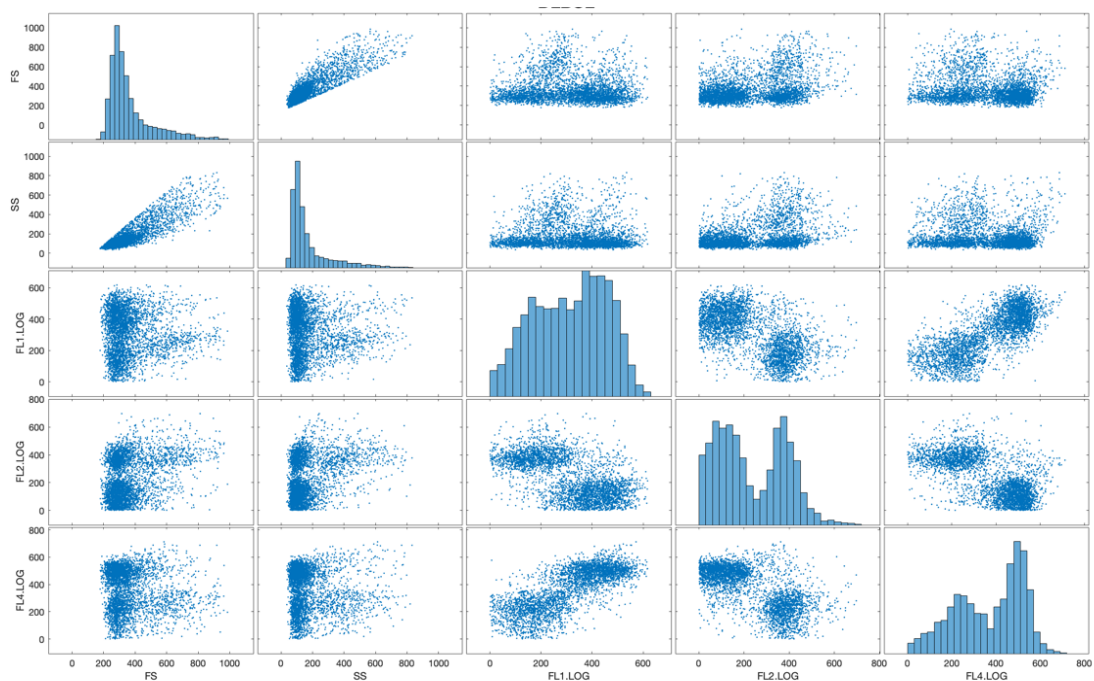
Figure 31. Scatter plot of DLBCL dataset

## 4.1.2 Symptomatic West Nile Virus (WNV)

The WNV (West Nile virus infection) dataset contains 13 samples from patients with WNV. Figure 32 is a scatter plot of this dataset. All samples contain above 87000 events
This dataset is provided by McMaster University[102] (Appendices 2: Details of Real Datasets, Table 33).

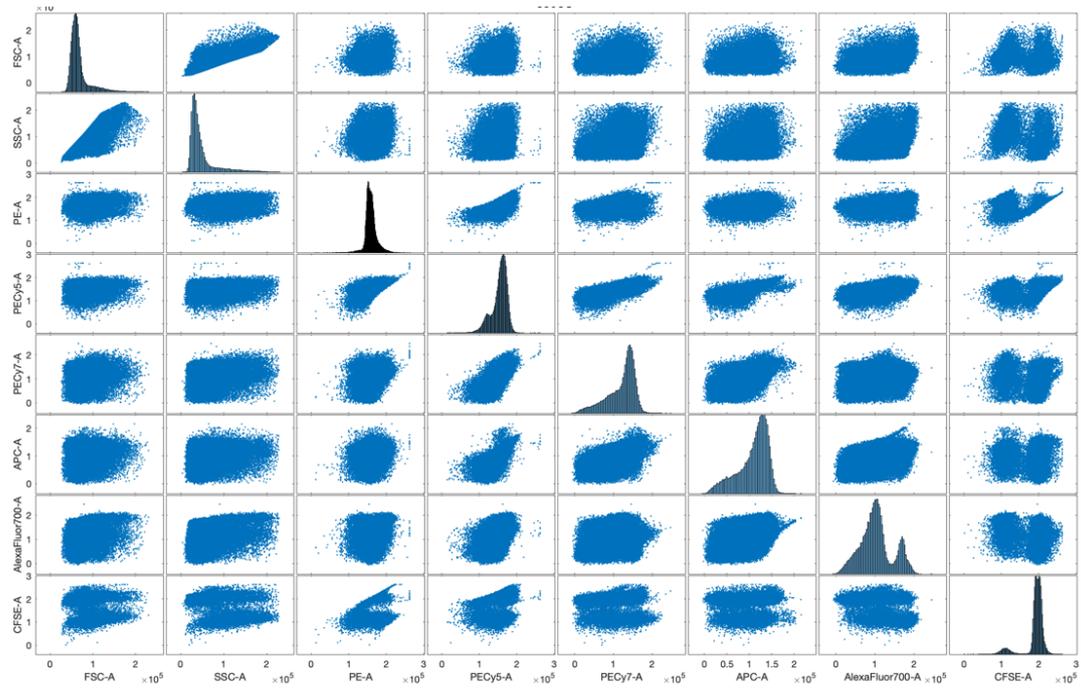
Figure 32. Scatter plot of WNV dataset

### 4.1.3 Hematopoietic Stem Cell Transplant (HSCT)

The HSCT dataset have 30 samples and was accessible by BCCRC[102]. Most samples have around 9000 cells. The scatter plot of HSCT dataset is presented in Figure 33, and the details of the dataset is available in Appendices 2: Details of Real Datasets, Table 32.

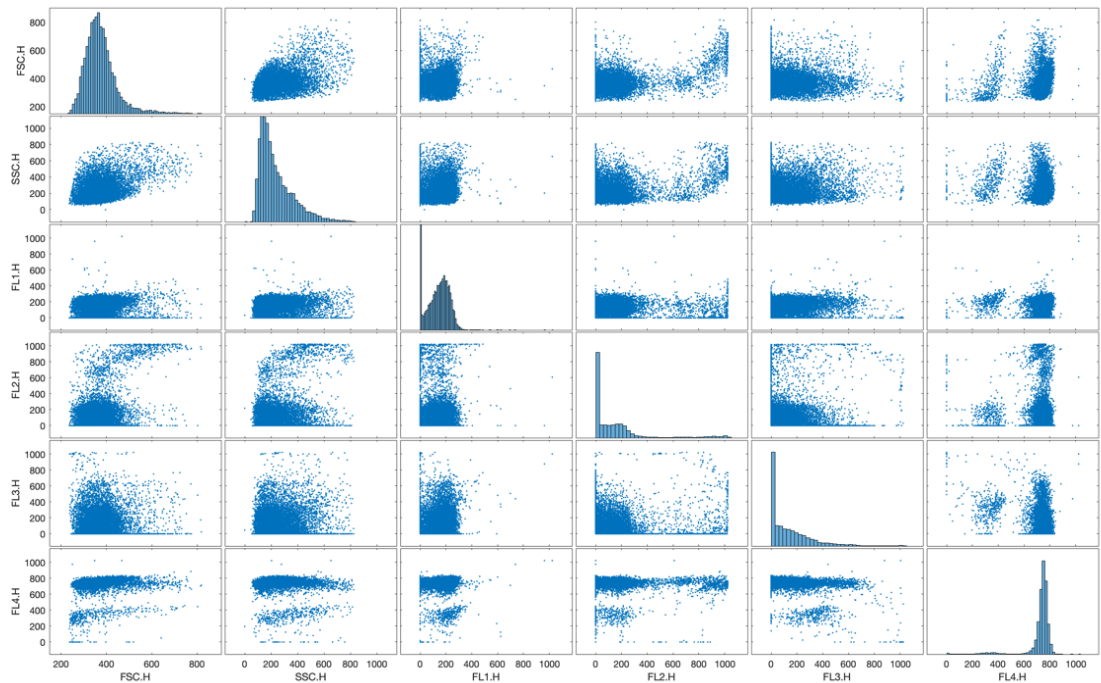
Figure 33. Scatter plot of HSCT dataset

## 4.1.4 Graft versus Host Disease (GvHD)

This dataset contains 12 samples from patients with GvHD (Graft-versus-Host disease) [103]. Cells stained with four markers: CD4, CD8b, CD3 and CD8. All samples in this dataset contain more than 12000 cells. The scatterplot of GvHD dataset is shown in Figure 34, and further details can be viewed at Appendices 2: Details of Real Datasets, Table 31.

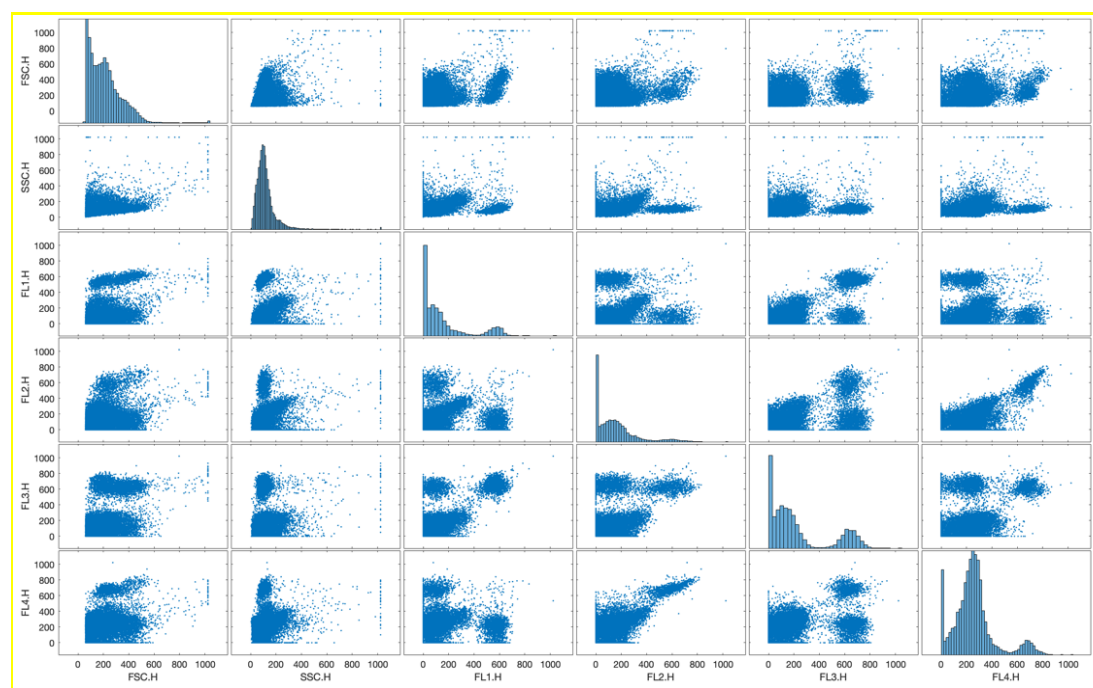

Figure 34. Scatter plot of GvHD dataset

The summary details of real datasets can be seen in Table 6.

Table 6. Real Datasets

| Dataset | # Samples | # Events | Marker |
|---|---|---|---|
| GvHD | 12 | 14,000 | CD4, CD8b, CD3, CD8 |
| DLBCL | 30 | 5,000 | CD3, CD5, CD19 |
| HSCT | 30 | 10,000 | CD45.1, Ly65/Mac1, CD45.2, Dead Cells |
| WNV | 13 | 100,000 | IFNg, CD3, CD4, IL17, CD8, Free Amines |

## 4.2 Synthetic Datasets

Real datasets are too complicated and sometimes hard to discover the potential shortcomings of the algorithm. Thus, it is necessary to use simple but efficient synthetic dataset to find the potential drawbacks of the algorithm. Two popular and effective synthetic datasets are used to break the algorithms: mouse sets and Jain sets.

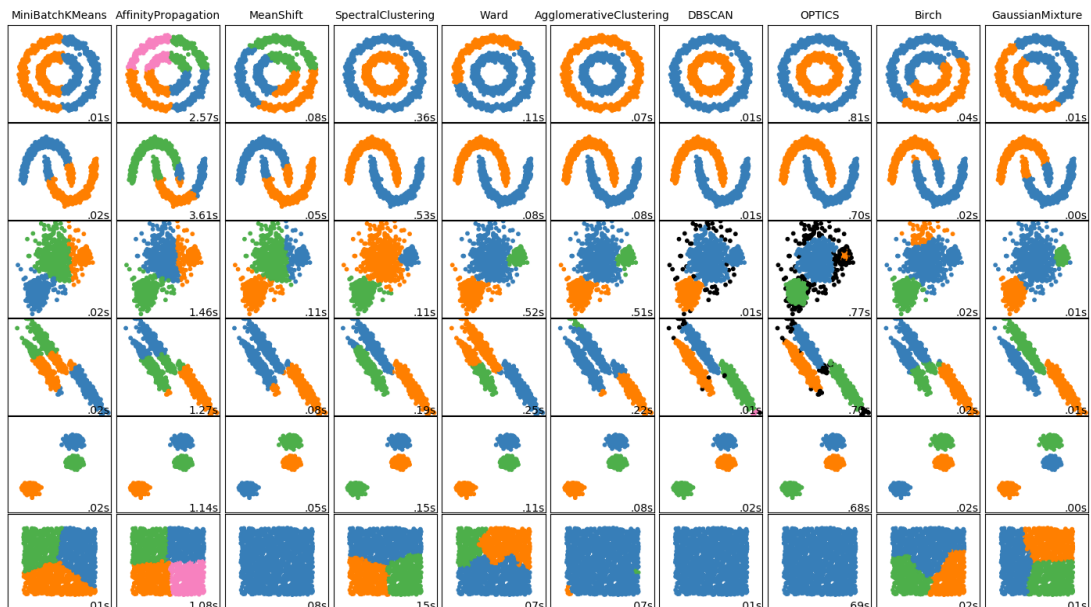

Figure 35. Comparing different clustering algorithms on toy datasets[104].

## 4.2.1 Mouse set

The Mouse set is a two-dimensional dataset that contains three Gaussian clusters and noise. The shape of mouse sets is similar to Mickey Mouse, the cartoon character. This dataset can be generated by ELKI [105]. The properties of mouse sets can be seen in Table 7, and Figure 36 shows a visualization of mouse set.

Table 7. Details of Mouse Set

| Attribute | Distribution | Mean (Var 1) | Standard dev (Var 1) | Mean (Var 2) | Standard dev (Var 2) | Size |
|---|---|---|---|---|---|---|
| Head | Gaussian | 0.5 | 0.2 | 0.5 | 0.2 | 290 |
| Left Ear | Gaussian | 0.25 | 0.05 | 0.75 | 0.05 | 100 |
| Right Ear | Gaussian | 0.75 | 0.05 | 0.75 | 0.05 | 100 |
| Noise | Uniform | | | | | 10 |

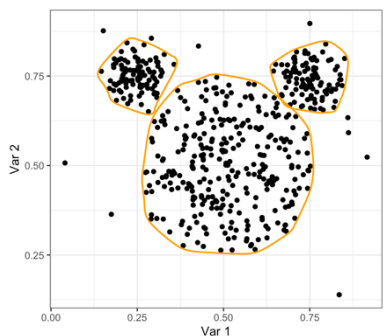
(a) Manual Gating

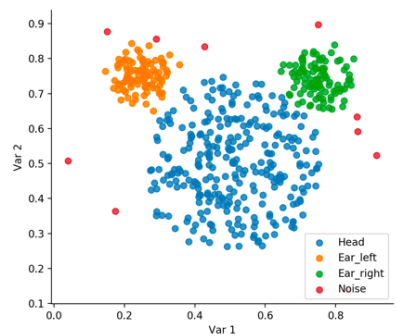
(b) Scatter plot

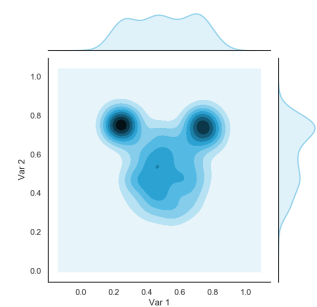
(c) Contour plot

Figure 36. Mouse set

### 4.2.2 Jain set

Jain set is a two-dimensional dataset from [106] and also be used as a benchmark dataset to discover the k-means properties [107]. This dataset can be acquired from [107]. The visualization of this dataset as shown in Figure 37.

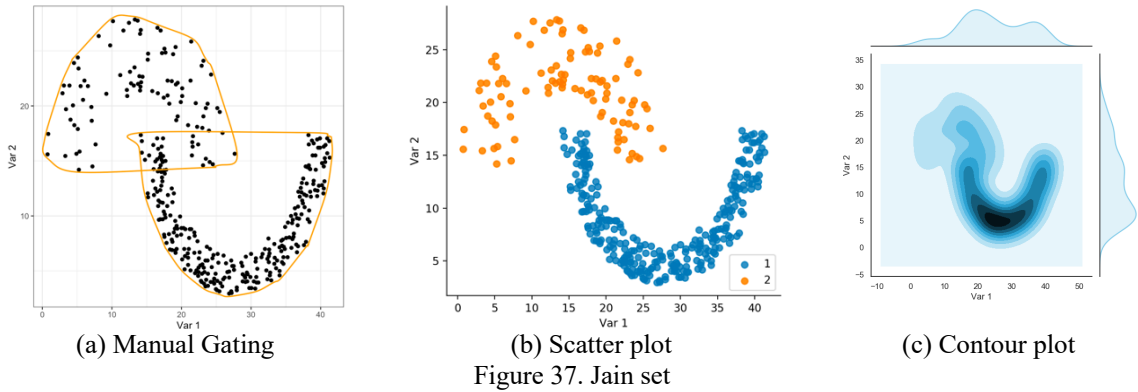

(a) Manual Gating     (b) Scatter plot     (c) Contour plot

Figure 37. Jain set

The size, dimension and number of clusters of synthetic datasets are illustrated in Table 8.

Table 8. Synthetic Datasets

| Dataset | Size | Dim | Number of clusters |
| --- | --- | --- | --- |
| Mouse | 500 | 2 | 3 |
| Jain | 373 | 2 | 2 |

## 4.3 Data Pre-processing

As previously mentioned in Section 2.3.2, steps of flow cytometry data pre-processing include compensation, transformation, quality control and normalisation. Therefore, the real data need to be compensated, transformed to linear space, and have irrelevant or sensitive information (e.g., patient information) as well as dead cells removed. These processes can be performed by FlowJo [50].

## 4.4 Evaluation Measurements

A popular basic evaluation tool is confusion matrix [54]. Figure 38 shows the structure of confusion matrix. The correct predictions are shown in green and the incorrect predictions are shown in red.

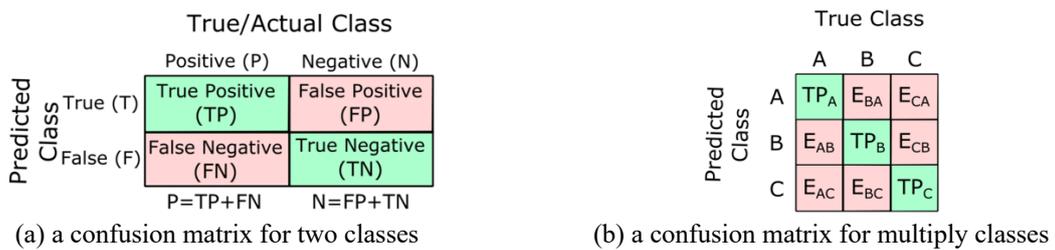

(a) a confusion matrix for two classes     (b) a confusion matrix for multiply classes

Figure 38. Confusion matrix [108]

The meanings of the four counts in the confusion matrix are as follows:
- True Positive (TP): the prediction is positive and true label is positive
- True Negative (TN): the prediction is negative and true label is negative
- False Positive (FP): the prediction is positive while true label is negative
- False Negative (FN): the prediction is negative while true label is positive

By tracking these metrics, a more detailed analysis of the performance of the model beyond the basic percentage of guesses that were correct can be achieved [54]. There are four basic evaluation measurements based on the confusion matrix: accuracy, precision, recall, and F1 score. This paper mainly uses the last three evaluations to analyse the performance of algorithms.

Precision (positive prediction value) represents the proportion of true classes to the total number of positive predictions. This term is indicated as follows [109]:

$$precision = \frac{TP}{TP + FP} = \frac{\text{number of cells correctly assigned to a cluster}}{\text{total cells allocated to that cluster}}$$

Recall (ture positive pate or sensitivity) represents the proportion of positive predictions to the total number of true classes and it can be defined as follows [109]:

$$recall = \frac{TP}{TP + FN} = \frac{\text{the number of cells correctly assigned to a cluster}}{\text{all the cells that should have been allocated to that cluster}}$$

A more intuitive explanation of precision and recall is provided in Figure 39.

F1 score (F1 measurement) represents the harmonic mean of precision and recall as indicated by the following equation [109]:

$$F_1 = 2 \times \frac{precision \times recall}{precision + recall} = \frac{2TP}{2TP + FP + FN}$$

Beyond that, a new measurement called correctness defined in this paper:

$$correctness = \frac{\text{Number of samples correctly identified number of clusters}}{\text{Total number of samples}}$$

If a sample has four clusters, and the algorithm also cluster the data to four groups, then the correctness equals to 1, otherwise the correctness equals to 0.

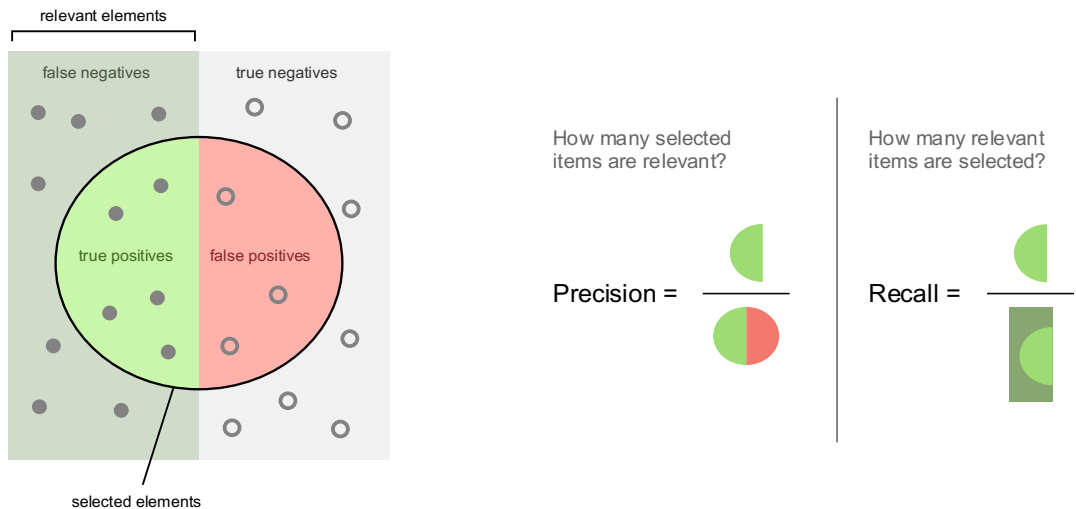

Figure 39. Precision and Recall [110]

The values of $F_1$ ranges from 0 to 1, with 1 meaning all predicted classes match true classes [102]. For multi-class tasks, the $F_1$ score can be calculated from the average of each class's $F_1$ score and weighted by the number of true instances for each class. In this analysis, the true class is the result of manual gating by human experts.

# 5 Comparison of Automatic Gating Algorithms

Current automated gating algorithms are mainly design for identify cell populations or classify the samples[102]. This thesis compares three representative automated gating algorithms developed for solving the problem of cell population identification: flowMeans, flowMerge and SamSPECTRAL.

## 5.1 Automatic Gating Algorithms

### 5.1.1 flowMerge

FlowMerge[89] is an extended version of flowClust [41], a model-based automated gating algorithm. The characteristics of flow cytometry data are mentioned in Section 2.2.2.1. Flow cytometry could be well represented by a mixture of distributions. However, the presentation of raw flow cytometry data may contain much noise, and cell populations tend to be asymmetric and overlapping. Thus, the Gaussian mixtures model is not enough to analyse flow cytometry data. To address such shortcomings, flowClust implements the Box-Cox transformation as part of the model fitting process, as the Box-Cox transformation allows for asymmetric populations. In addition, flowClust uses multivariate-t mixtures to enhance robustness and the Bayesian information criterion (BIC) to estimate the number of clusters. However, flowClust tends to use multiple mixture components to represent the same cluster, as shown in Figure 40.

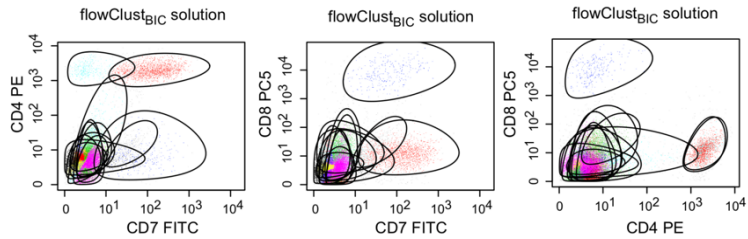
Figure 40.

To address such problem, flowMerge [89] merges overlapping mixture components based on entropy, and grouped merged components using a single multivariate–t distribution based on moment matching conditions. Thus, flowMerge can give a better estimate of the number of cell subpopulations than flowClust, as shown in Figure 41. The implementation of flowMerge is available in Bioconductor [111].

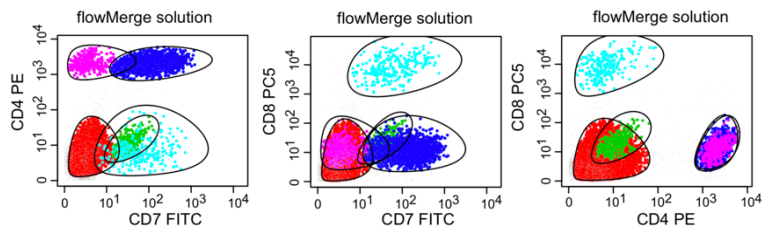
Figure 41.

### 5.1.2 flowMeans

The first automated clustering algorithm applied for flow cytometry data is based on K-means [18]. The traditional K-means based algorithm is faster than others, but it heavily relies on pre-defined numbers of clusters and cannot cluster spherical clusters well [71]. Consequently, more complexity models were developed to overcome such limitations, such as Gaussian mixture models [112], t mixture models [41] and skew-t mixture models [88]. These complexity models are better able to capture the complexity of flow cytometry; however, they need more time than the traditional K-means based algorithm. FlowMeans is an expansion of flowMerge [89] that uses a simple K-means model to decrease the complexity of the algorithm.

The main idea of flowMeans is to use overfitting to overcome the limitations of the K-means algorithm. The biggest limitation of the traditional K-means algorithm is that it cannot assign numbers of clusters itself. Therefore, flowMeans uses the total number of modes as that maximal number of clusters initially and then, merges the two closest clusters based on Mahalanobis distance until only one cluster is left. Finally, flowMeans uses a segmented regression algorithm to find the change-point in which the clusters are well separated. These steps are described in Figure 42. The implementation of flowMeans is available through Bioconductor [113].

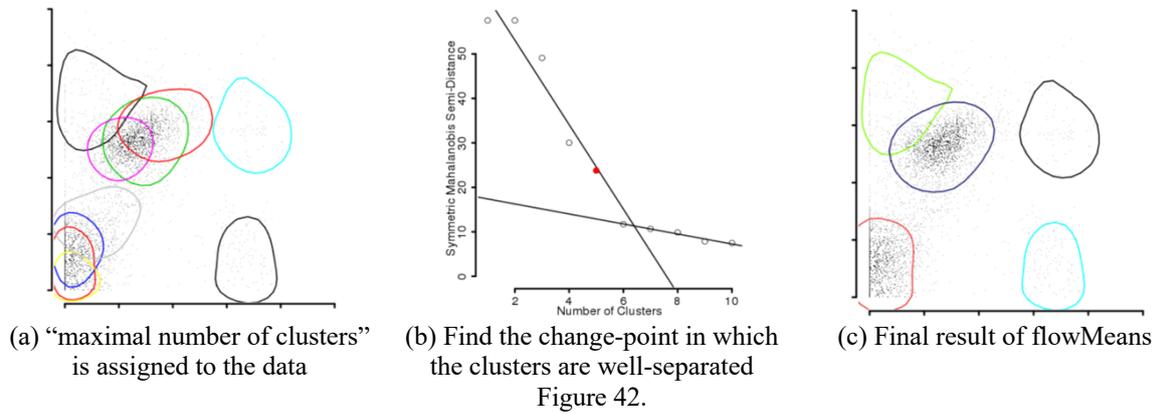

(a) "maximal number of clusters" is assigned to the data
(b) Find the change-point in which the clusters are well-separated
(c) Final result of flowMeans

Figure 42.

### 5.1.3 SamSPECTRAL

SamSPECTRAL uses modified spectral clustering to identify cell populations, especially bas better performance, in identifying rare cell populations [95]. Spectral clustering is a non-parametric clustering algorithm that can make up for the shortcomings of parametric clustering algorithms. Thus, spectral clustering does not need to know or assume the shape, size or density of clusters first. In addition, spectral clustering is robust to the noise and shape of clusters. However, one major limitation of spectral clustering is its huge demand for time and memory, similar to the model-based automated gating algorithm, which makes it unsuitable to being used for flow cytometry data directly. Thus, SamSPECTRAL uses a non-uniform information preserving sampling procedure to reduce the dimension of data (data reduction scheme) [95], which makes SamSPECTRAL analyse high-dimensional flow cytometry data efficiently. The major processes of the data reduction scheme are shown in Figure 43. First, SamSPECTRAL uses a representative subsampling method called faithful sampling to reduce the number of the graph's vertices created by spectral clustering (Figure 43 b), then a similarity matrix is used to assign the higher weights to edges (Figure 43 c).

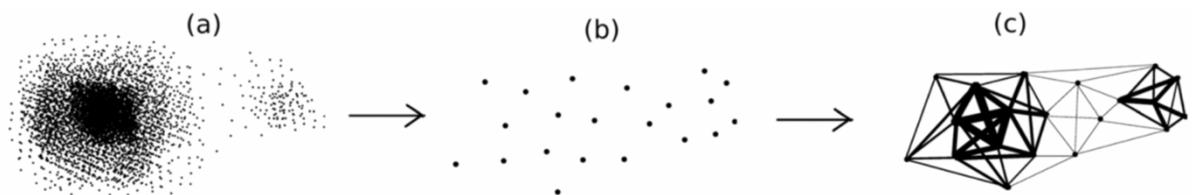

(a) Running spectral clustering is impractical on data that contains thousands of points. (b) Faithful sampling picks up a reasonable subset of points such that running spectral clustering is possible on them. However, all information about the local density is lost by considering only these sample points. (c) We assign weights to the edges of the graph; the edges between the nodes in denser regions are weighted considerably higher. The information about the local density is retrieved in this way.

Figure 43. Data Reduction Scheme of SamSPECTRAL

The pseudocode faithful sampling algorithm used in the data reduction scheme is defined by the following:

1. Label all data points as *unregistered*.
2. **repeat**
3.     Pick a random unregistered point *p* {the representative of a new community}
4.     Label all unregistered data points within distance *h* from *p* as *registering*
5.     Put registering points in a set called *community p*
6.     Relabel registering points as *registered*
7. **until** All points are registered
8. **return** All communities

## 5.2 Results

This thesis evaluates the performance of algorithms based on their completely automated ability and partial automated ability:
- Completely automated ability: algorithms not assigned with a predefined number of clusters.
- Partial automated ability: algorithms assigned with a predefined number of clusters.

All required parameters are fixed before implemented and used across all datasets.

### 5.2.1 Synthetic Datasets

During the evaluation process, the required parameters of flowMerge and SamSPECTRAL are set to commonly used values in advance and used for both completely automated and partial automated processes; the details can be seen in Table 9.

Table 9. The Setting of Required Parameters

| Required parameters | Mouse set | Jain set | Descriptions |
|---|---|---|---|
| SamSPECTRAL | | | |
| normal.sigma | 200 | | The scaling parameter, the larger it is the algorithm will find smaller clusters. |
| separation.factor | 0.7 | | This threshold controls to what extend clusters should be combined or kept separate. |
| flowMerge | | | |
| level | 0.8 | 0.9 | A numeric value between 0 and 1 specifying the threshold quantile level used to call a point an outlier. The default is 0.9, meaning that any point outside the 90% quantile region will be called an outlier. |

According to Figure 44 and Table 11, it is clear that flowMeans has less completely automated ability in synthetic datasets than others, SamSPRCTRSL overestimates the number of clusters for both the mouse set and the Jain set, and flowMerge seems to perform well in the mouse set.

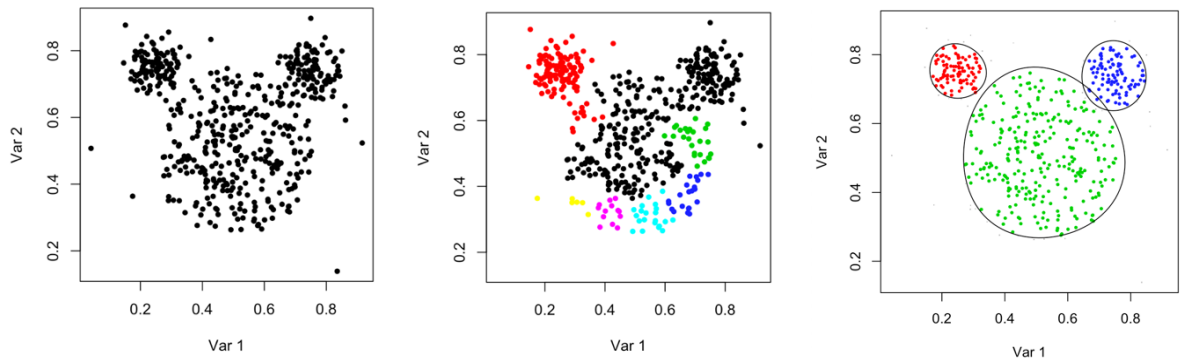

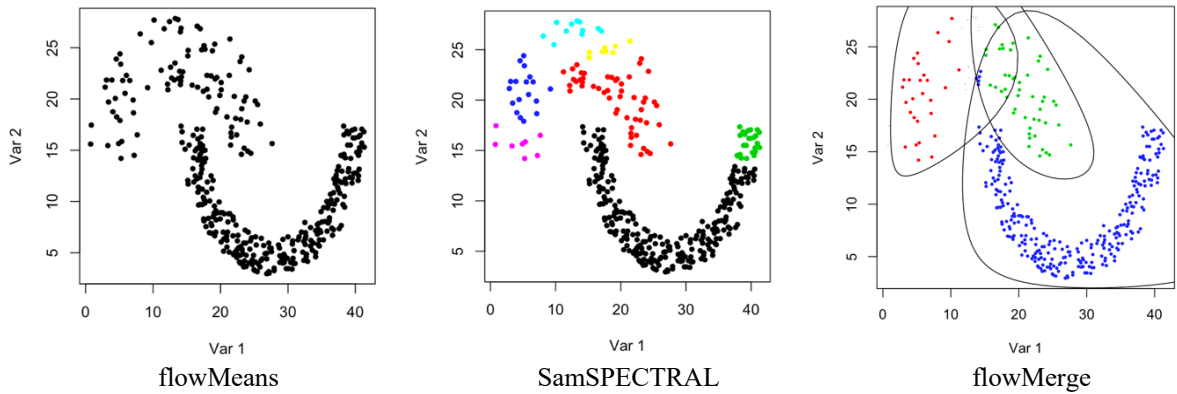

flowMeans            SamSPECTRAL            flowMerge

Figure 44. Completely Automated Results

The completely automated results are summarised in Table 10. The overall completely automated performance of flowMerge is higher than the others. Estimating the correct number of clusters is not the major task of SamSPECTRAL. Therefore, SamSPECTRAL still performs well in the Jain set even though it estimates that the number of clusters in the Jain set equals 7. FlowMerge correctly estimates the number of clusters in the mouse set and identifies nearly all clusters correctly.

Table 10. Completely Automated

|  | $F_1$ score (Precision, Recall) | | Mean |
|---|---|---|---|
|  | **Mouse** | **Jain** |  |
| flowMeans | 0.067 (0.04, 0.2) | 0.629 (0.548, 0.739) | 0.348 |
| SamSPECTRAL | 0.576 (0.573, 0.592) | **0.885** (1, 0.809) | 0.730 |
| flowMerge | **0.988** (0.988, 0.988) | 0.830 (0.954, 0.809) | **0.909** |

Table 11. Completely Automated

|  | # Clusters | |
|---|---|---|
|  | **Mouse** | **Jain** |
| flowMeans | 1 | 1 |
| SamSPECTRAL | 7 | 7 |
| flowMerge | 3 | 3 |

Figure 45 shows a visualization results of flowMeans, flowMerge and SamSPECTRAL.

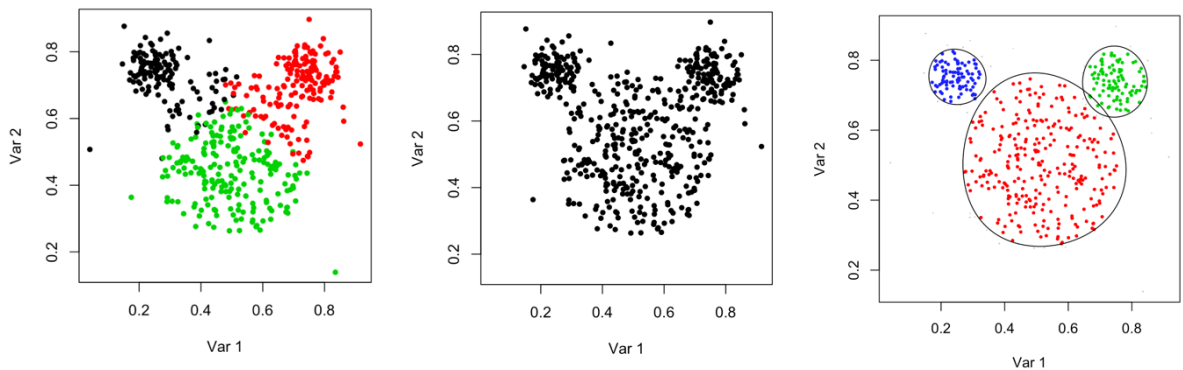

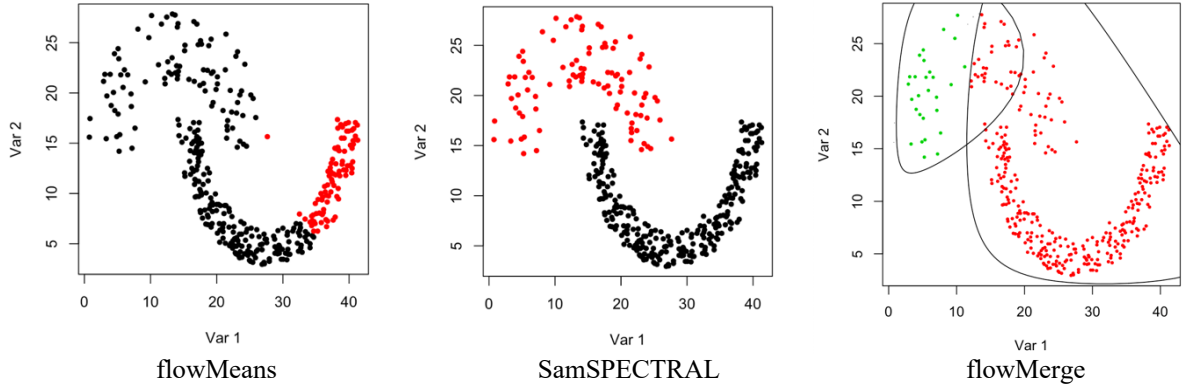

| flowMeans | SamSPECTRAL | flowMerge |

Figure 45. Partial Automated

As shown in Table 12, after assigning the correct number of clusters, the performance of all algorithms is improved, especially flowMeans. FlowMerge still has the best performance. SamSPECTRAL can identify all clusters in the Jain set perfectly, but overall performance is reduced. Assigning the predefined or correct number of clusters cannot guarantee that the algorithm will achieve better performance. The performance of flowMeans and flowMerge in the Jain set are reduced, and SamSPECTRAL also achieves a lower $F_1$ score in the mouse set compared with completely automated.

Table 12. Partial Automated

|  | $F_1$ score (Precision, Recall) | | Mean |
|---|---|---|---|
|  | **Mouse** | **Jain** |  |
| flowMeans | 0.793 (0.844, 0.8) | 0.486 (0.828, 0.493) | 0.639 |
| SamSPECTRAL | 0.433 (0.357, 0.584) | **1** (1, 1) | 0.716 |
| flowMerge | **0.988** (0.988, 0.988) | 0.773 (0.854, 0.809) | **0.994** |

### 5.2.2 Real Datasets

The required parameters of SamSPECTRAL and flowMerge also fixed in advance and used for all real datasets, as shown in Table 13.

Table 13. The Setting of Required Parameters

| **Required parameters** | **Real Datasets** | **Descriptions** |
|---|---|---|
| | | SamSPECTRAL |
| normal.sigma | 200 | The scaling parameter, the larger it is the algorithm will find smaller clusters. |
| separation.factor | 0.7 | This threshold controls to what extend clusters should be combined or kept separate. |
| | | flowMerge |
| level | 0.8 | A numeric value between 0 and 1 specifying the threshold quantile level used to call a point an outlier. The default is 0.9, meaning that any point outside the 90% quantile region will be called an outlier. |

During completely automated processing, SamSPECTRAL achieves better performance in DLBCL, GvHD and HSCT datasets, the overall performance is also better than other algorithms, as shown in Table 14.

Table 14. Completely Automated

|  | $F_1$ score (Precision, Recall) | | | | Mean |
|---|---|---|---|---|---|
|  | **DLBCL** | **GvHD** | **HSCT** | **WNV** |  |
| flowMeans | 0.820 | 0.721 | 0.864 | **0.760** | 0.791 |

|  | (0.829, 0.828) | (0.773, 0.703) | (0.907, 0.832) | (0.811, 0.726) |  |
|---|---|---|---|---|---|
| SamSPECTRAL | **0.860** | **0.792** | **0.918** | 0.702 | **0.818** |
|  | (0.910, 0.843) | (0.854, 0.777) | (0.965, 0.893) | (0.777, 0.682) |  |
| flowMerge | 0.837 | 0.446 | 0.501 | 0.725 | 0.627 |
|  | (0.896, 0.803) | (0.775, 0.341) | (0.912, 0.361) | (0.908, 0.635) |  |

However, SamSPECTRAL has poor ability to correctly estimate the number of clusters for all datasets, as presented in Table 15. The possible reason why SamSPECTRAL cannot estimate the number of clusters correctly but also can achieve better performance is the purpose of SamSPECTRAL is not for correctly estimate the number of clusters, but for determine the rare cell populations. [95]

Table 15. Completely Automated

|  | Correctness | | | | Mean |
|---|---|---|---|---|---|
|  | DLBCL | GvHD | HSCT | WNV |  |
| flowMeans | 0.26 | 0 | 0.033 | 0.230 | 0.130 |
| SamSPECTRAL | 0 | 0 | 0 | 0 | 0 |
| flowMerge | 0.26 | 0 | 0.033 | 0.230 | 0.130 |

The results for partial automated gating are shown in Table 16. The $F_1$ score of flowMeans in DLBCL and GvHD increased a little bit, while the $F_1$ score of others in all real datasets are decreased. Thus, even if the algorithm cannot correctly estimate the number of clusters, it still cannot guarantee that assign a pre-defined number of clusters will improve the performance of algorithms.

Table 16. Partial Automated

|  | $F_1$ score (Precision, Recall) | | | | Mean |
|---|---|---|---|---|---|
|  | DLBCL | GvHD | HSCT | WNV |  |
| flowMeans | **0.852** | **0.725** | **0.838** | **0.713** | **0.782** |
|  | (0.866, 0.856) | (0.675, 0.799) | (0.819, 0.874) | (0.707, 0.743) |  |
| SamSPECTRAL | 0.622 | 0.646 | 0.597 | 0.542 | 0.601 |
|  | (0.648, 0.707) | (0.611, 0.741) | (0.583, 0.697) | (0.485, 0.671) |  |
| flowMerge | 0.804 | 0.430 | 0.411 | 0.604 | 0.562 |
|  | (0.817, 0.837) | (0.588, 0.387) | (0.719, 0.315) | (0.551, 0.702) |  |

### 5.2.2.1 Further Discussion

To deeper explore the performance of flowMeans, flowMerge and SamSPECTRAL in real datasets, some characteristic samples was selected for further discussion. The selected samples are all huge unbalanced (the difference between the least number clusters and the most numerous clusters up to 100 times more), the details can be seen in Table 17.

Table 17. Selected Characteristic Real Data

| Samples | # Events | # Cluster 1 | # Cluster 2 | # Cluster 3 | # Cluster 4 |
|---|---|---|---|---|---|
| DLBCL #28 | 21011 | 20674 | 183 |  |  |
| GvHD #11 | 32700 | 873 | 31174 | 67 |  |
| GvHD #12 | 16336 | 1402 | 14300 | 86 |  |
| HSCT #1 | 9936 | 14 | 397 | 8621 | 754 |
| HSCT #3 | 9790 | 9135 | 505 | 47 |  |
| HSCT #4 | 9694 | 570 | 8856 | 128 | 37 |
| HSCT #29 | 7210 | 5906 | 69 | 361 | 21 |
| WNV #1 | 92498 | 5586 | 543 | 11861 | 67265 |
| WNV #11 | 94489 | 1417 | 337 | 68690 | 16932 |
| WNV #12 | 90036 | 16517 | 71294 | 413 |  |

According to Table 18 and Table 19, SamSPECTRAL did not has better overall performance, but it performs well in most huge unbalanced data, while flowMerge is not suitable for analysing such types of data.

Table 18. Complete Automated

|  | DLBCL #28 | GvHD #11 | GvHD #12 | HSCT #1 | HSCT #3 | HSCT #4 | HSCT #29 | WNV #1 | WNV #11 | WNV #12 |
|---|---|---|---|---|---|---|---|---|---|---|
| **flowMeans** | **0.976** | 0.84 | 0.952 | 0.936 | 0.863 | 0.935 | 0.754 | **0.793** | **0.845** | 0.818 |
| **flowMerge** | 0.921 | 0.383 | 0.329 | 0.478 | 0.543 | 0.498 | 0.613 | 0.741 | 0.784 | 0.27 |
| **SamSPECTRAL** | 0.966 | **0.978** | **0.962** | **0.942** | **0.936** | **0.943** | **0.822** | 0.703 | 0.825 | **0.866** |

Table 19. Partial Automated

|  | DLBCL #28 | GvHD #11 | GvHD #12 | HSCT #1 | HSCT #3 | HSCT #4 | HSCT #29 | WNV #1 | WNV #11 | WNV #12 |
|---|---|---|---|---|---|---|---|---|---|---|
| **flowMeans** | 0.698 | **0.972** | **0.951** | 0.732 | **0.974** | **0.97** | 0.691 | 0.589 | **0.82** | **0.852** |
| **flowMerge** | **0.983** | 0.386 | 0.332 | 0.303 | 0.543 | 0.498 | 0.529 | **0.7** | 0.637 | 0.703 |
| **SamSPECTRAL** | 0.978 | 0.971 | 0.825 | 0.385 | 0.963 | 0.888 | **0.758** | 0.614 | 0.614 | 0.702 |

# 6 Subspace Clustering

The research of clustering analysis focuses on scalability, the effectiveness of clustering complexity data, high-dimensional data and mixed data. Among them, high-dimensional data clustering is the difficulty problem in clustering analysis. However, traditional clustering algorithms encounter difficulties in clustering high-dimensional data spaces. To solve this problem, R. Agrawal first proposed the concept of subspace clustering [115].

The subspace clustering algorithm extends the task of feature selection and attempts to discover clusters on different subspaces of the same dataset. As with feature selection, subspace clustering requires the use of a search strategy and evaluation criteria to select groups that need to be clustered. However, considering that different clusters exist in different subspaces, some limitations on the evaluation criteria are required.

The selected search strategy has a large impact on the clustering results. According to the direction of the search, the subspace clustering method can be divided into two categories: top-down search strategy and bottom-up search strategy, as shown in Figure 46.

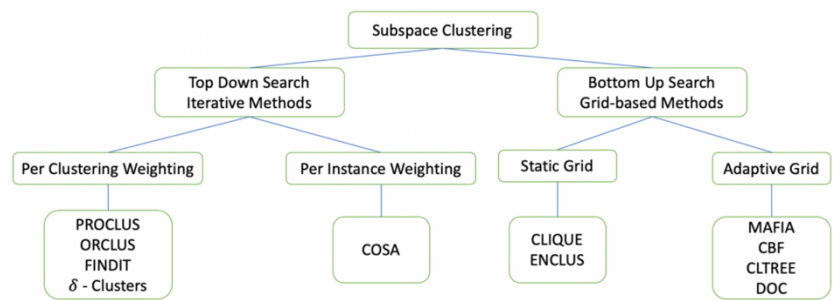

At the top there are two main types of subspace clustering algorithms, distinguished by their search strategy. One group uses a top-down approach that iteratively updates weights for dimensions for each cluster. The other uses a bottom-up approach, building clusters up from dense units found in low dimensional space. Each of those groups is then further categorized based on how they define a context for the quality measure.

Figure 46. Hierarchy of subspace clustering [114]

## 6.1 ProClus Algorithm

ProClus [116] uses a top-down strategy that performs in a same way as K-medoids. ProClus mainly has three stages: a medoids that keep away from each other are first chosen by greed algorithm in initialization stage, k medoids will be randomly selected and replace the bad medoids. The working flow of ProClus is shown in Figure 47.

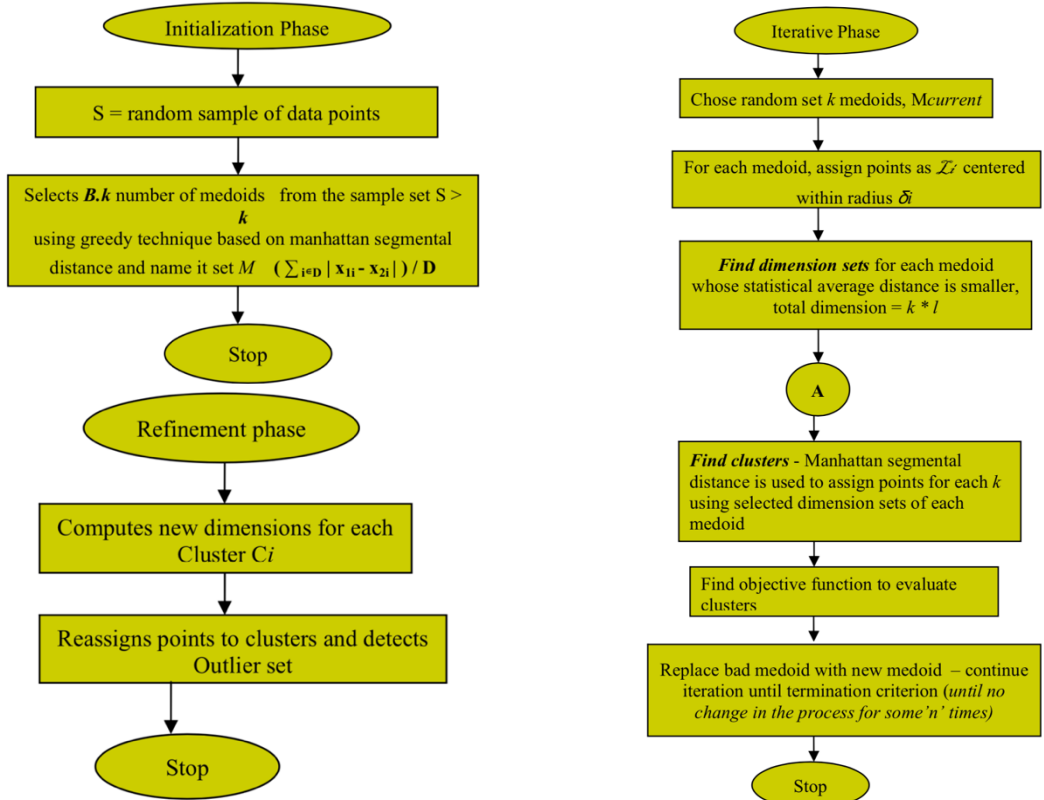

Figure 47. Flow chart of ProClus running phrase [117]

## 6.2 Results

As Table 20 shows, ProClus has poor ability to analyse flow cytometry data for both completely and partial automated gating. The performance of ProClus is improved a bit after assigning the pre-defined number of clusters. The correctness of ProClus seems not too bad, but ProClus prone to divide all datasets into four clusters. Most real data used in this thesis are have four clusters, that the reason why ProClus has higher correctness than flowMeans, flowMerge and SamSPECTRAL.

Table 20. $F_1$ Score of Real Datasets

| | $F_1$ score (Precision, Recall) | | | | Mean |
|---|---|---|---|---|---|
| | **DLBCL** | **GvHD** | **HSCT** | **WNV** | |
| Completely Automated | 0.515 (0.803, 0.439) | 0.417 (0.771, 0.328) | 0.447 (0.735, 0.356) | 0.415 (0.664, 0.337) | 0.448 |
| Partial Automated | 0.564 (0.738, 0.504) | 0.609 (0.738, 0.571) | 0.474 (0.689, 0.407) | 0.481 (0.586, 0.455) | 0.532 |
| Correctness (Completely Automated) | 0.2 | 0.5 | 0.366 | 0.923 | 0.497 |

Table 21 presents ProClus's results of synthetic datasets. ProClus has slight better performance in synthetic dataset than in real dataset. As the same as other three automated gating algorithms, assign the pre-defined number of clusters cannot always improve the performance of ProClus.

Table 21. F$_1$ Score of Synthetic Datasets

|  | F$_1$ score (Precision, Recall) | | | # Clusters |
|---|---|---|---|---|
|  | **Mouse** | **Jain** | **Mean** |  |
| Completely Automated | 0.448 (0.915, 0.432) | 0.580 (0.905, 0.461) | 0.514 | Mouse: 4; Jain: 4 |
| Partial Automated | 0.470 (0.509, 0.446) | 0.512 (0.533, 0.493) | 0.491 |  |

# 7 Comparison of ProClus and Automated Gating Algorithms

When evaluated by a completely automated process in real datasets, ProClus cannot beat any of the three algorithms; it can achieve a better performance than flowMeans only in the Jain set, as shown in Table 22.

Table 22. Completely automated

|  | Real Datasets | | | | Synthetic Datasets | |
|---|---|---|---|---|---|---|
|  | **DLBCL** | **GvHD** | **HSCT** | **WNV** | **Mouse** | **Jain** |
| flowMeans | 0.820 | 0.721 | 0.864 | 0.760 | 0.067 | 0.486 |
| SamSPECTRAL | 0.860 | 0.792 | 0.918 | 0.702 | 0.576 | 1 |
| flowMerge | 0.837 | 0.446 | 0.501 | 0.725 | 0.988 | 0.773 |
| ProClus | 0.515 | 0.417 | 0.447 | 0.415 | 0.448 | **0.580** |

As shown in Table 23, after assigning the predefined number of clusters to ProClus, its performance in real datasets improves and has slightly better performance than flowMerge in the GvHD and HSCT datasets. In synthetic datasets, the performance of ProClus in the mouse set also decreases, like SamSPECTRAL.

Table 23. Partial Automated

|  | Real Datasets | | | | Synthetic Datasets | |
|---|---|---|---|---|---|---|
|  | **DLBCL** | **GvHD** | **HSCT** | **WNV** | **Mouse** | **Jain** |
| flowMeans | 0.852 | 0.725 | 0.838 | 0.713 | 0.793 | 0.629 |
| SamSPECTRAL | 0.622 | 0.646 | 0.597 | 0.542 | 0.433 | 0.885 |
| flowMerge | 0.804 | 0.430 | 0.411 | 0.604 | 0.988 | 0.830 |
| ProClus | 0.564 | **0.609** | **0.474** | 0.481 | **0.470** | 0.512 |

The overall performance of ProClus is lower than the other three existing automated gating algorithms. However, ProClus has a better ability to ignore outliers (uninterested cells) than others. Combining the results of Table 24 and Table 25, ProClus has a better ability to ignore uninterested cells than other algorithms for both completely automated and partially automated processes, and this may be the reason why ProClus achieves a bad performance (it treats many cells as uninterested).

Table 24. Completely Automated

|  | **DLBCL** | **GvHD** | **HSCT** | **WNV** | **Mean** |
|---|---|---|---|---|---|
| flowMeans | 0 | 0 | 0 | 0 | 0 |
| SamSPECTRAL | 0.0010 | 0.0013 | 0.0023 | 0.0005 | 0.0012 |
| flowMerge | 0.0047 | 0.0140 | 0.0114 | 0.0028 | 0.0082 |
| flowMeans | 0 | 0 | 0 | 0 | 0 |
| SamSPECTRAL | 0.0011 | 0.0013 | 0.0023 | 0.0005 | 0.0012 |
| flowMerge | 0.0047 | 0.0140 | 0.0114 | 0.0028 | 0.0082 |
| ProClus | **0.0216** | **0.0174** | **0.0097** | **0.0243** | **0.0183** |

# 8 Conclusion and Future Work

In this thesis, the machine learning techniques used in the flow cytometry data analysis are discussed. A conventional flow cytometry data analysis called 'manual gating' heavily relies on experts' experience and is manually performed. Therefore, the conventional analysis method for flow cytometry data is highly subjective, time-consuming, labour-wasting and unsuitable for flow cytometry. To overcome the limitations of manual gating, many machine-learning-based gating methods are developed, especially unsupervised-learning-based gating methods. In current unsupervised machine-learning algorithms, clustering-based algorithms are the most used in flow cytometry data analysis.

The three commonly used clustering-based automated algorithms, flowMeans, flowMerge and SanSPECTRUL, are first applied, compared and evaluated by both complete and partial automated processes in four real datasets and two synthetic datasets. Another subspace-clustering-based algorithm, ProClus, that is not used in flow cytometry areas, participated in. According to the evaluation results, the existing automated gating algorithms still cannot achieve completely automated; some of them still need to be assigned the pre-defined number of clusters. However, some performances of existing automated algorithms will decrease after assigning the pre-defined number of clusters. The reason for this phenomenon may be that the purpose of the algorithm design is different. Some automated gating algorithms are designed to identify rare cell populations; a correct estimate for the number of clusters is not the goal in such algorithms as SamSPECTRAL.

In future work, more automated gating algorithms should be evaluated and compared. New methods to evaluate the algorithms' performances still need to be discovered because correct manual gating results from humans cannot be guaranteed.

# Reference

[1] Errante, P., Ebbing, P., Rodrigues, F., Ferraz, R. and Da Silva, N. (2019). *Flow cytometry: a literature review*.

[2] Advances in ME/CFS. (2019). *Approaches in Biomedical Research: Flow Cytometry, Part 2*. [online] Available at: https://jaxmecfs.com/2018/02/23/approaches-in-biomedical-research-flow-cytometry-part-2/

[3] Robinson, J.P., 2004. Flow cytometry. *Encyclopedia of biomaterials and biomedical engineering*, *3*, pp.630-642.

[4] Bio-rad-antibodies.com. (2019). [online] Available at: https://www.bio-rad-antibodies.com/static/2017/flow/flow-cytometry-basics-guide.pdf

[5] M. Brown and C. Wittwer. Flow cytometry: Principles and clinical applications in hematology. *Clinical Chemistry*, 46:1221–1229, 2000.

[6] Lou, X., Zhang, G., Herrera, I., Kinach, R., Ornatsky, O., Baranov, V., Nitz, M. and Winnik, M. (2007). Polymer-Based Elemental Tags for Sensitive Bioassays. *Angewandte Chemie*, 119(32), pp.6223-6226.

[7] Majonis, D., Herrera, I., Ornatsky, O., Schulze, M., Lou, X., Soleimani, M., Nitz, M. and Winnik, M. (2010). Synthesis of a Functional Metal-Chelating Polymer and Steps toward Quantitative Mass Cytometry Bioassays. *Analytical Chemistry*, 82(21), pp.8961-8969.

[8] Barteneva, N.S., Fasler-Kan, E. and Vorobjev, I.A., 2012. Imaging flow cytometry: coping with heterogeneity in biological systems. *Journal of Histochemistry & Cytochemistry*, *60*(10), pp.723-733.

[9] Spidlen, J., Breuer, K., Rosenberg, C., Kotecha, N. and Brinkman, R.R., 2012. FlowRepository: A resource of annotated flow cytometry datasets associated with peer-reviewed publications. *Cytometry Part A*, *81*(9), pp.727-731.

[10] Bioconductor.org. (2019). *Bioconductor - BiocViews*. [online] Available at: http://www.bioconductor.org/packages/release/BiocViews.html#___FlowCytometry

[11] O'Neill, K., Aghaeepour, N., Špidlen, J. and Brinkman, R., 2013. Flow cytometry bioinformatics. *PLoS computational biology*, *9*(12), p.e1003365.

[12] Reich, M., Liefeld, T., Gould, J., Lerner, J., Tamayo, P. and Mesirov, J.P., 2006. GenePattern 2.0. *Nature genetics*, *38*(5), p.500

[13] Software.broadinstitute.org. (2019). *GenePattern*. [online] Available at: https://software.broadinstitute.org/cancer/software/genepattern/genepattern-flow-cytometry-suite.

[14] Datascienceguide.github.io. (2019). *Exploratory Data Analysis*. [online] Available at: https://datascienceguide.github.io/exploratory-data-analysis

[15] Inspiration Information. (2019). *Audio Signals in Python*. [online] Available at: http://myinspirationinformation.com/uncategorized/audio-signals-in-python/

[16] Medium. (2019). *Time Series Analysis 1*. [online] Available at: https://towardsdatascience.com/time-series-analysis-1-9f4360f43110.

[17] Medium. (2019). *Analyze the data through data visualization using Seaborn*. [online] Available at: https://towardsdatascience.com/analyze-the-data-through-data-visualization-using-seaborn-255e1cd3948e.

[18] Murphy, R.F., 1985. Automated identification of subpopulations in flow cytometric list mode data using cluster analysis. *Cytometry: The Journal of the International Society for Analytical Cytology*, *6*(4), pp.302-309.

[19] O'Neill, K., Jalali, A., Aghaeepour, N., Hoos, H. and Brinkman, R.R., 2014. Enhanced flowType/RchyOptimyx: a bioconductor pipeline for discovery in high-dimensional cytometry data. *Bioinformatics*, *30*(9), pp.1329-1330.

[20] Qi, X. and Gao, X., 2015. Towards a better understanding of mouse and human diseases—International Mouse Phenotyping Consortium. *Science China Life Sciences*, *58*(4), pp.392-395.

[21] Murphy, R.F. and Chused, T.M., 1984. A proposal for a flow cytometric data file standard. *Cytometry: The Journal of the International Society for Analytical Cytology*, *5*(5), pp.553-555.

[22] Isac-net.org. (2019). *International Society for Advancement of Cytometry*. [online] Available at: https://isac-net.org/default.aspx

[23] Dean, P.N., Bagwell, C.B., Lindmo, T., Murphy, R.F. and Salzman, G.C., 1990. Introduction to flow cytometry data file standard. *Cytometry: The Journal of the International Society for Analytical Cytology*, *11*(3), pp.321-322.

[24] Seamer, L.C., Bagwell, C.B., Barden, L., Redelman, D., Salzman, G.C., Wood, J.C.S. and Murphy, R.F., 1997. Proposed new data file standard for flow cytometry, version FCS 3.0. *Cytometry: The Journal of the International Society for Analytical Cytology*, *28*(2), pp.118-122.

[25] Spidlen, J., Moore, W., Parks, D., Goldberg, M., Bray, C., Bierre, P., Gorombey, P., Hyun, B., Hubbard, M., Lange, S. and Lefebvre, R., 2010. Data file standard for flow cytometry, version FCS 3.1. *Cytometry Part A: The Journal of the International Society for Advancement of Cytometry*, *77*(1), pp.97-100.

[26] Spidlen, Josef & Shooshtari, Parisa & Kollmann, Tobias & Brinkman, Ryan. (2011). Flow cytometry data standards. *BMC research notes*. 4. 50. 10.1186/1756-0500-4-50.

[27] Lee, J.A., Spidlen, J., Boyce, K., Cai, J., Crosbie, N., Dalphin, M., Furlong, J., Gasparetto, M., Goldberg, M., Goralczyk, E.M. and Hyun, B., 2008. MIFlowCyt: the minimum information about a Flow Cytometry Experiment. *Cytometry Part A: The journal of the International Society for Analytical Cytology*, 73(10), pp.926-930

[28] Spidlen, J., Breuer, K. and Brinkman, R., 2012. Preparing a Minimum Information about a Flow Cytometry Experiment (MIFlowCyt) compliant manuscript using the International Society for Advancement of Cytometry (ISAC) FCS file repository (FlowRepository. org). *Current protocols in cytometry*, 61(1), pp.10-18.

[29] Spidlen, J., Leif, R.C., Moore, W., Roederer, M. and Brinkman, R.R., 2008. Gating-ML: XML-based gating descriptions in flow cytometry. *Cytometry Part A: The Journal of the International Society for Analytical Cytology*, 73(12), pp.1151-1157.

[30] Maecker, H.T., Rinfret, A., D'Souza, P., Darden, J., Roig, E., Landry, C., Hayes, P., Birungi, J., Anzala, O., Garcia, M. and Harari, A., 2005. Standardization of cytokine flow cytometry assays. *BMC immunology*, 6(1), p.13.

[31] Spidlen, J., Bray, C., ISAC Data Standards Task Force and Brinkman, R.R., 2015. ISAC's classification results file format. *Cytometry Part A*, 87(1), pp.86-88.

[32] Chattopadhyay, P.K., Gierahn, T.M., Roederer, M. and Love, J.C., 2014. Single-cell technologies for monitoring immune systems. *Nature immunology*, 15(2), p.128.

[33] Bashashati, A. and Brinkman, R.R., 2009. A survey of flow cytometry data analysis methods. *Advances in bioinformatics*, 2009.

[34] Tung, J.W., Heydari, K., Tirouvanziam, R., Sahaf, B., Parks, D.R., Herzenberg, L.A. and Herzenberg, L.A., 2007. Modern flow cytometry: a practical approach. *Clinics in laboratory medicine*, 27(3), pp.453-468.

[35] Roederer, M., 2002. Compensation in flow cytometry. *Current protocols in cytometry*, 22(1), pp.1-14.

[36] Hahne, F., LeMeur, N., Brinkman, R.R., Ellis, B., Haaland, P., Sarkar, D., Spidlen, J., Strain, E. and Gentleman, R., 2009. flowCore: a Bioconductor package for high throughput flow cytometry. *BMC bioinformatics*, 10(1), p.106.

[37] Shapiro, H.M., 2005. *Practical flow cytometry*. John Wiley & Sons.

[38] Parks, D.R., Roederer, M. and Moore, W.A., 2006. A new "Logicle" display method avoids deceptive effects of logarithmic scaling for low signals and compensated data. *Cytometry Part A: The Journal of the International Society for Analytical Cytology*, 69(6), pp.541-551.

[39] Moore, W.A. and Parks, D.R., 2012. Update for the logicle data scale including operational code implementations. *Cytometry Part A*, 81(4), pp.273-277.

[40] Bagwell, C.B., 2005. Hyperlog—A flexible log-like transform for negative, zero, and positive valued data. *Cytometry Part A: The Journal of the International Society for Analytical Cytology*, 64(1), pp.34-42.

[41] Lo, K., Brinkman, R.R. and Gottardo, R., 2008. Automated gating of flow cytometry data via robust model-based clustering. *Cytometry Part A: the journal of the International Society for Analytical Cytology*, 73(4), pp.321-332.

[42] Le Meur, N., Rossini, A., Gasparetto, M., Smith, C., Brinkman, R.R. and Gentleman, R., 2007. Data quality assessment of ungated flow cytometry data in high throughput experiments. *Cytometry Part A: The Journal of the International Society for Analytical Cytology*, 71(6), pp.393-403.

[43] Rogers, W.T., Moser, A.R., Holyst, H.A., Bantly, A., Mohler III, E.R., Scangas, G. and Moore, J.S., 2008. Cytometric fingerprinting: quantitative characterization of multivariate distributions. *Cytometry Part A: The Journal of the International Society for Analytical Cytology*, 73(5), pp.430-441.

[44] Hastie, T., Tibshirani, R., Friedman, J. and Franklin, J., 2005. The elements of statistical learning: data mining, inference and prediction. *The Mathematical Intelligencer*, 27(2), pp.83-85.

[45] Lugli, E., Roederer, M. and Cossarizza, A., 2010. Data analysis in flow cytometry: the future just started. *Cytometry Part A*, 77(7), pp.705-713.

[46] Khowawisetsuta, L., Sukapirom, K. and Pattanapanyasatb, K., 2018. Data analysis and presentation in flow cytometry. *SCIENCEASIA*, 44, pp.19-27.

[47] Brunk, C.F., Bohman, R.E. and Brunk, C.A., 1982. Conversion of linear histogram flow cytometry data to a logarithmic display. *Cytometry: The Journal of the International Society for Analytical Cytology*, 3(2), pp.138-141.

[48] Finak, G., Perez, J.M., Weng, A. and Gottardo, R., 2010. Optimizing transformations for automated, high throughput analysis of flow cytometry data. *BMC bioinformatics*, 11(1), p.546.

[49] Herzenberg, L.A., Tung, J., Moore, W.A., Herzenberg, L.A. and Parks, D.R., 2006. Interpreting flow cytometry data: a guide for the perplexed. *Nature immunology*, 7(7), p.681.

[50] Flowjo.com. (2019). *FlowJo | FlowJo, LLC*. [online] Available at: https://www.flowjo.com/solutions/flowjo.
[51] De Novo Software. (2019). *Homepage*. [online] Available at: https://denovosoftware.com
[52] Antibodies-online.com. (2019). *What is flow cytometry (FACS analysis)?*. [online] Available at: https://www.antibodies-online.com/resources/17/1247/what-is-flow-cytometry-facs-analysis/
[53] Montante, S. and Brinkman, R.R., 2019. Flow cytometry data analysis: Recent tools and algorithms. *International journal of laboratory hematology*, *41*, pp.56-62.
[54] Patterson, J. and Gibson, A., 2017. *Deep learning: A practitioner's approach*. " O'Reilly Media, Inc.".
[55] Medium. (2019). *Different types of Machine learning and their types.*. [online] Available at: https://medium.com/deep-math-machine-learning-ai/different-types-of-machine-learning-and-their-types-34760b9128a2.
[56] Malek, M., Taghiyar, M.J., Chong, L., Finak, G., Gottardo, R. and Brinkman, R.R., 2014. flowDensity: reproducing manual gating of flow cytometry data by automated density-based cell population identification. *Bioinformatics*, *31*(4), pp.606-607.
[57] Johnson, S.C., 1967. Hierarchical clustering schemes. *Psychometrika*, *32*(3), pp.241-254.
[58] Murtagh, F., 1983. A survey of recent advances in hierarchical clustering algorithms. *The Computer Journal*, *26*(4), pp.354-359.
[59] Uc-r.github.io. (2019). *Hierarchical Cluster Analysis · UC Business Analytics R Programming Guide*. [online] Available at: https://uc-r.github.io/hc_clustering.
[60] Zhang, T., Ramakrishnan, R. and Livny, M., 1996, June. BIRCH: an efficient data clustering method for very large databases. In *ACM Sigmod Record* (Vol. 25, No. 2, pp. 103-114). ACM.
[61] ROCK, G.S.R.K., A robust clustering algorithm for categorical attributes. *Sydney⋅ Proceedings of the 15th ICDE'1999*, pp.512-521.
[62] Guha, S., Rastogi, R. and Shim, K., 1998, June. CURE: an efficient clustering algorithm for large databases. In *ACM Sigmod Record* (Vol. 27, No. 2, pp. 73-84). ACM.
[63] Adinis.sk. (2019). *Products / ADICyt — Adinis Ltd.*. [online] Available at: http://www.adinis.sk/en/products/bioinformatics-and-data-processing/adicyt.html
[64] Casanas, A. (2019). *Hierarchical Clustering - Coding with Alex*. [online] Coding with Alex. Available at: https://codingwithalex.com/hierarchical-clustering/
[65] Murphy, K.P., 2012. *Machine learning: a probabilistic perspective*. MIT press.
[66] Park, H.S. and Jun, C.H., 2009. A simple and fast algorithm for K-medoids clustering. *Expert systems with applications*, *36*(2), pp.3336-3341.
[67] MacQueen, J., 1967, June. Some methods for classification and analysis of multivariate observations. In *Proceedings of the fifth Berkeley symposium on mathematical statistics and probability* (Vol. 1, No. 14, pp. 281-297).
[68] Ng, R.T. and Han, J., 2002. CLARANS: A method for clustering objects for spatial data mining. *IEEE Transactions on Knowledge & Data Engineering*, (5), pp.1003-1016.
[69] Kaufman, L. and Rousseeuw, P.J., 1990. Partitioning around medoids (program pam). *Finding groups in data: an introduction to cluster analysis*, *344*, pp.68-125.
[70] Kaufman, L. and Rousseeuw, P.J., 2009. *Finding groups in data: an introduction to cluster analysis* (Vol. 344). John Wiley & Sons.
[71] Aghaeepour, N., Nikolic, R., Hoos, H.H. and Brinkman, R.R., 2011. Rapid cell population identification in flow cytometry data. *Cytometry Part A*, *79*(1), pp.6-13.
[72] RDSG. (2019). *K-Means Clustering: Explained*. [online] Available at: https://iitrdsg.wordpress.com/2016/06/15/k-means-clustering-explained/.
[73] Kriegel, H.P., Kröger, P., Sander, J. and Zimek, A., 2011. Density-based clustering. *Wiley Interdisciplinary Reviews: Data Mining and Knowledge Discovery*, *1*(3), pp.231-240.
[74] Ester, M., Kriegel, H.P., Sander, J. and Xu, X., 1996, August. A density-based algorithm for discovering clusters in large spatial databases with noise. In *Kdd*(Vol. 96, No. 34, pp. 226-231).
[75] Ankerst, M., Breunig, M.M., Kriegel, H.P. and Sander, J., 1999, June. OPTICS: ordering points to identify the clustering structure. In *ACM Sigmod record* (Vol. 28, No. 2, pp. 49-60). ACM.
[76] Comaniciu, D. and Meer, P., 2002. Mean shift: A robust approach toward feature space analysis. *IEEE Transactions on Pattern Analysis & Machine Intelligence*, (5), pp.603-619.
[77] Naumann, U., Luta, G. and Wand, M.P., 2010. The curvHDR method for gating flow cytometry samples. *BMC bioinformatics*, *11*(1), p.44.
[78] Zaki, M.J., 2001. SPADE: An efficient algorithm for mining frequent sequences. *Machine learning*, *42*(1-2), pp.31-60.
[79] Sugar, I.P. and Sealfon, S.C., 2010. Misty Mountain clustering: application to fast unsupervised flow cytometry gating. *BMC bioinformatics*, *11*(1), p.502.

[80] Jhui.github.io. (2019). *"Machine learning - Clustering, Density based clustering and SOM"*. [online] Available at: https://jhui.github.io/2017/01/15/Machine-learning-clustering/.
[81] Han, J., Pei, J. and Kamber, M., 2011. *Data mining: concepts and techniques*. Elsevier.
[82] Agrawal, R., Gehrke, J., Gunopulos, D. and Raghavan, P., 1998. *Automatic subspace clustering of high dimensional data for data mining applications* (Vol. 27, No. 2, pp. 94-105). ACM.
[83] Qian, Y., Wei, C., Eun-Hyung Lee, F., Campbell, J., Halliley, J., Lee, J.A., Cai, J., Kong, Y.M., Sadat, E., Thomson, E. and Dunn, P., 2010. Elucidation of seventeen human peripheral blood B-cell subsets and quantification of the tetanus response using a density-based method for the automated identification of cell populations in multidimensional flow cytometry data. *Cytometry Part B: Clinical Cytometry*, *78*(S1), pp.S69-S82.
[84] Sharma, A., Gupta, R.K. and Tiwari, A., 2016. Improved Density Based Spatial Clustering of Applications of Noise Clustering Algorithm for Knowledge Discovery in Spatial Data. *Mathematical Problems in Engineering*, *2016*.
[85] Rasmussen, C.E., 2000. The infinite Gaussian mixture model. In *Advances in neural information processing systems* (pp. 554-560).
[86] Fisher, D.H., 1987. Knowledge acquisition via incremental conceptual clustering. *Machine learning*, *2*(2), pp.139-172.
[87] Chan, C., Feng, F., Ottinger, J., Foster, D., West, M. and Kepler, T.B., 2008. Statistical mixture modeling for cell subtype identification in flow cytometry. *Cytometry Part A: The Journal of the International Society for Analytical Cytology*, *73*(8), pp.693-701.
[88] Pyne, S., Hu, X., Wang, K., Rossin, E., Lin, T.I., Maier, L.M., Baecher-Allan, C., McLachlan, G.J., Tamayo, P., Hafler, D.A. and De Jager, P.L., 2009. Automated high-dimensional flow cytometric data analysis. *Proceedings of the National Academy of Sciences*, *106*(21), pp.8519-8524.
[89] Finak, G., Bashashati, A., Brinkman, R. and Gottardo, R., 2009. Merging mixture components for cell population identification in flow cytometry. *Advances in bioinformatics*, *2009*.
[90] Xu, D. and Tian, Y., 2015. A comprehensive survey of clustering algorithms. *Annals of Data Science*, *2*(2), pp.165-193.
[91] Lapan, M., 2018. *Deep Reinforcement Learning Hands-On: Apply modern RL methods, with deep Q-networks, value iteration, policy gradients, TRPO, AlphaGo Zero and more*. Packt Publishing Ltd.
[92] Li, H., Shaham, U., Stanton, K.P., Yao, Y., Montgomery, R.R. and Kluger, Y., 2017. Gating mass cytometry data by deep learning. *Bioinformatics*, *33*(21), pp.3423-3430.
[93] Eulenberg, P., Köhler, N., Blasi, T., Filby, A., Carpenter, A.E., Rees, P., Theis, F.J. and Wolf, F.A., 2017. Reconstructing cell cycle and disease progression using deep learning. *Nature communications*, *8*(1), p.463.
[94] Doan, M., Sebastian, J.A., Pinto, R.N., McQuin, C., Goodman, A., Wolkenhauer, O., Parsons, M.J., Acker, J.P., Rees, P., Hennig, H. and Kolios, M.C., 2018. Label-free assessment of red blood cell storage lesions by deep learning. *BioRxiv*, p.256180.
[95] Zare, H., Shooshtari, P., Gupta, A. and Brinkman, R.R., 2010. Data reduction for spectral clustering to analyze high throughput flow cytometry data. *BMC bioinformatics*, *11*(1), p.403.
[96] El Khettabi, F. and Kyriakidis, P., 2006. The L 2 discrepancy framework to mine high-throughput screening data for targeted drug discovery: application to AIDS antiviral activity data of the National Cancer Institute. In *Data Mining for Biomedical Informatics workshop, SIAM Conf. Data Mining*.
[97] Quinn, J., Fisher, P.W., Capocasale, R.J., Achuthanandam, R., Kam, M., Bugelski, P.J. and Hrebien, L., 2007. A statistical pattern recognition approach for determining cellular viability and lineage phenotype in cultured cells and murine bone marrow. *Cytometry Part A: The Journal of the International Society for Analytical Cytology*, *71*(8), pp.612-624.
[98] Naim, I., Datta, S., Sharma, G., Cavenaugh, J.S. and Mosmann, T.R., 2010, March. Swift: scalable weighted iterative sampling for flow cytometry clustering. In *2010 IEEE International Conference on Acoustics, Speech and Signal Processing* (pp. 509-512). IEEE.
[99] Ge, Y. and Sealfon, S.C., 2012. flowPeaks: a fast unsupervised clustering for flow cytometry data via K-means and density peak finding. *Bioinformatics*, *28*(15), pp.2052-2058.
[100] Aghaeepour, N., Chattopadhyay, P.K., Ganesan, A., O'Neill, K., Zare, H., Jalali, A., Hoos, H.H., Roederer, M. and Brinkman, R.R., 2012. Early immunologic correlates of HIV protection can be identified from computational analysis of complex multivariate T-cell flow cytometry assays. *Bioinformatics*, *28*(7), pp.1009-1016.
[101] Qian, Y., Wei, C., Eun-Hyung Lee, F., Campbell, J., Halliley, J., Lee, J.A., Cai, J., Kong, Y.M., Sadat, E., Thomson, E. and Dunn, P., 2010. Elucidation of seventeen human peripheral blood B-cell subsets and quantification of the tetanus response using a density-based method for the automated identification of cell populations in multidimensional flow cytometry data. *Cytometry Part B: Clinical Cytometry*, *78*(S1), pp.S69-S82.

[102] Aghaeepour, N., Finak, G., Hoos, H., Mosmann, T.R., Brinkman, R., Gottardo, R., Scheuermann, R.H., FlowCAP Consortium and Dream Consortium, 2013. Critical assessment of automated flow cytometry data analysis techniques. *Nature methods*, *10*(3), p.228.
[103] Brinkman, R.R., Gasparetto, M., Lee, S.J.J., Ribickas, A.J., Perkins, J., Janssen, W., Smiley, R. and Smith, C., 2007. High-content flow cytometry and temporal data analysis for defining a cellular signature of graft-versus-host disease. *Biology of Blood and Marrow Transplantation*, *13*(6), pp.691-700.
[104] Scikit-learn.org. (2019). *Comparing different clustering algorithms on toy datasets — scikit-learn 0.21.3 documentation*. [online] Available at: https://scikit-learn.org/stable/auto_examples/cluster/plot_cluster_comparison.html
[105] Elki-project.github.io. (2019). *ELKI Data Mining Framework*. [online] Available at: https://elki-project.github.io.
[106] Jain, A.K. and Law, M.H., 2005, December. Data clustering: A user's dilemma. In *International conference on pattern recognition and machine intelligence* (pp. 1-10). Springer, Berlin, Heidelberg.
[107] Cs.joensuu.fi. (2019). *Clustering datasets*. [online] Available at: http://cs.joensuu.fi/sipu/datasets/.
[108] Tharwat, A. (2018). Classification assessment methods. *Applied Computing and Informatics*.
[109] Sokolova, M., Japkowicz, N. and Szpakowicz, S., 2006, December. Beyond accuracy, F-score and ROC: a family of discriminant measures for performance evaluation. In *Australasian joint conference on artificial intelligence* (pp. 1015-1021). Springer, Berlin, Heidelberg.
[110] En.wikipedia.org. (2019). *F1 score*. [online] Available at: https://en.wikipedia.org/wiki/F1_score
[111] Bioconductor. (2019). *flowClust*. [online] Available at: http://www.bioconductor.org/packages/release/bioc/html/flowClust.html.
[112] Chan, C., Feng, F., Ottinger, J., Foster, D., West, M. and Kepler, T.B., 2008. Statistical mixture modeling for cell subtype identification in flow cytometry. *Cytometry Part A: The Journal of the International Society for Analytical Cytology*, *73*(8), pp.693-701.
[113] Bioconductor.org. (2019). *Bioconductor - flowMeans*. [online] Available at: http://www.bioconductor.org/packages/2.6/bioc/html/flowMeans.html.
[114] Parsons, L., Haque, E. and Liu, H., 2004. Subspace clustering for high dimensional data: a review. *Acm Sigkdd Explorations Newsletter*, *6*(1), pp.90-105.
[115] Agrawal, R., Gehrke, J., Gunopulos, D. and Raghavan, P., 1998. *Automatic subspace clustering of high dimensional data for data mining applications* (Vol. 27, No. 2, pp. 94-105). ACM.
[116] Aggarwal, C.C., Wolf, J.L., Yu, P.S., Procopiuc, C. and Park, J.S., 1999, June. Fast algorithms for projected clustering. In *ACM SIGMoD Record* (Vol. 28, No. 2, pp. 61-72). ACM.
[117] Nayagam, S.C., Comparative Study of Subspace Clustering Algorithms.

# Appendices 1: Detailed Results Per Sample

Table 26. DLBCL

| # | flowMeans | | flowMerge | | SamSPECTRAL | | ProClus | |
|---|---|---|---|---|---|---|---|---|
| | Completely automated | Partial Automated | Completely automated | Partial Automated | Completely automated | Partial Automated | Completely automated | Partial Automated |
| | $F_1$ score (Precision, Recall) | | | | | | | |
| 1 | 0.723 (0.965,0.654) | 0.978 (0.975, 0.982) | 0.913 (0.983, 0.855) | 0.418 (0.373, 0.540) | 0.857 (0.990, 0.762) | 0.428 (0.340, 0.577) | 0.334 (0.751,0.224) | 0.831 (0.887,0.806) |
| 2 | 0.985 (0.981, 0.989) | 0.985 (0.981, 0.989) | 0.948 (0.987, 0.913) | 0.903 (0.936, 0.924) | 0.980 (0.985, 0.974) | 0.605 (0.855,0.4714) | 0.581 (0.946,0.444) | 0.598 (0.931,0.457) |
| 3 | 0.980 (0.976, 0.984) | 0.980 (0.976, 0.984) | 0.832 (0.877, 0.793) | 0.976 (0.985, 0.970) | 0.902 (0.989, 0.830) | 0.826 (0.779, 0.881) | 0.604 (0.949,0.451) | 0.563 (0.882,0.421) |
| 4 | 0.768 (0.779, 0.764) | 0.768 (0.779, 0.764) | 0.731 (0.842, 0.646) | 0.595 (0.647, 0.631) | 0.672 (0.855, 0.584) | 0.429 (0.424, 0.578) | 0.612 (0.646,0.584) | 0.621 (0.716,0.609) |
| 5 | 0.793 (0.872, 0.734) | 0.903 (0.883, 0.924) | 0.920 (0.916, 0.926) | 0.920 (0.916, 0.926) | 0.912 (0.913, 0.913) | 0.895 (0.876, 0.923) | 0.604 (0.864,0.532) | 0.704 (0.893,0.597) |
| 6 | 0.991 (0.990, 0.992) | 0.991 (0.990, 0.992) | 0.873 (0.997, 0.796) | 0.987 (0.995, 0.979) | 0.986 (0.995, 0.977) | 0.760 (0.698, 0.834) | 0.5 (0.99,0.365) | 0.544 (0.625,0.482) |
| 7 | 0.752 (0.697, 0.825) | 0.790 (0.746, 0.841) | 0.785 (0.824, 0.772) | 0.748 (0.758, 0.777) | 0.706 (0.834, 0.641) | 0.333 (0.283, 0.493) | 0.579 (0.782,0.486) | 0.533 (0.627,0.497) |
| 8 | 0.847 (0.968, 0.753) | 0.916 (0.916, 0.918) | 0.910 (0.974, 0.856) | 0.892 (0.910, 0.883) | 0.874 (0.911, 0.850) | 0.537 (0.773, 0.668) | 0.391 (0.907,0.345) | 0.515 (0.712,0.417) |
| 9 | 0.925 (0.945, 0.906) | 0.935 (0.922, 0.951) | 0.878 (0.959, 0.811) | 0.928 (0.926, 0.933) | 0.888 (0.963, 0.826) | 0.458 (0.429, 0.598) | 0.819 (0.937,0.759) | 0.708 (0.787,0.658) |
| 10 | 0.924 (0.901, 0.948) | 0.924 (0.901, 0.948) | 0.871 (0.931, 0.828) | 0.919 (0.926, 0.913) | 0.810 (0.952, 0.740) | 0.074 (0.562, 0.190) | 0.461 (0.938,0.376) | 0.492 (0.752,0.396) |
| 11 | 0.716 (0.650, 0.797) | 0.726 (0.684, 0.781) | 0.786 (0.826, 0.777) | 0.417 (0.418, 0.546) | 0.788 (0.834, 0.784) | 0.383 (0.341, 0.539) | 0.6 (0.82,0.538) | 0.698 (0.766,0.691) |
| 12 | 0.804 (0.821, 0.845) | 0.804 (0.821, 0.845) | 0.865 (0.901, 0.859) | 0.807 (0.864, 0.840) | 0.871 (0.907, 0.866) | 0.624 (0.614, 0.725) | 0.524 (0.487, 0.575) | 0.416 (0.643,0.358) |
| 13 | 0.610 (0.610, 0.611) | 0.737 (0.754, 0.786) | 0.732 (0.760, 0.777) | 0.732 (0.760, 0.777) | 0.709 (0.704, 0.759) | 0.708 (0.680, 0.778) | 0.337 (0.462,0.312) | 0.382 (0.681,0.37) |
| 14 | 0.9414 (0.930, 0.954) | 0.888 (0.918, 0.863) | 0.761 (0.967, 0.686) | 0.560 (0.763, 0.665) | 0.911 (0.944, 0.888) | 0.522 (0.457, 0.655) | 0.476 (0.956, 0.343) | 0.504 (0.721, 0.404) |
| 15 | 0.872 (0.915,0.838) | 0.906 (0.893,0.92) | 0.895 (0.931,0.864) | 0.915 (0.918,0.928) | 0.91 (0.905,0.92) | 0.665 (0.662,0.678) | 0.637 (0.875,0.519) | 0.485 (0.688,0.407) |
| 16 | 0.793 (0.737,0.858) | 0.905 (0.919,0.901) | 0.787 (0.858,0.729) | 0.862 (0.909,0.886) | 0.96 (0.979,0.946) | 0.804 (0.757,0.86) | 0.539 (0.862,0.406) | 0.6 (0.85,0.475) |
| 17 | 0.311 (0.301,0.387) | 0.712 (0.672,0.765) | 0.717 (0.744,0.748) | 0.566 (0.626,0.65) | 0.655 (0.761,0.722) | 0.522 (0.619,0.639) | 0.565 (0.632,0.558) | 0.466 (0.595,0.441) |
| 18 | 0.749 (0.712,0.792) | 0.896 (0.893,0.906) | 0.812 (0.878,0.759) | 0.703 (0.682,0.779) | 0.899 (0.924,0.895) | 0.852 (0.871,0.896) | 0.462 (0.796,0.356) | 0.736 (0.775,0.719) |
| 19 | 0.77 (0.71,0.843) | 0.733 (0.927,0.622) | 0.923 (0.971,0.88) | 0.964 (0.967,0.962) | 0.957 (0.966,0.959) | 0.775 (0.724,0.843) | 0.315 (0.841,0.212) | 0.433 (0.839,0.311) |
| 20 | 0.868 (0.889,0.84) | 0.954 (0.943,0.965) | 0.846 (0.899,0.802) | 0.907 (0.922,0.918) | 0.879 (0.902,0.867) | 0.952 (0.955,0.96) | 0.642 (0.859,0.568) | 0.151 (0.586,0.105) |
| 21 | 0.739 (0.709,0.798) | 0.722 (0.699,0.761) | 0.625 (0.663,0.591) | 0.935 (0.934,0.937) | 0.939 (0.948,0.938) | 0.539 (0.459,0.669) | 0.376 (0.74,0.344) | 0.759 (0.79,0.733) |
| 22 | 0.787 (0.73,0.854) | 0.714 (0.831,0.632) | 0.911 (0.98,0.87) | 0.867 (0.871,0.866) | 0.85 (0.849,0.854) | 0.79 (0.803,0.855) | 0.314 (0.872,0.201) | 0.899 (0.945,0.878) |
| 23 | 0.903 (0.934,0.883) | 0.937 (0.933,0.945) | 0.93 (0.963,0.909) | 0.826 (0.8,0.858) | 0.888 (0.897,0.893) | 0.786 (0.829,0.85) | 0.579 (0.737,0.49) | 0.047 (0.088,0.049) |
| 24 | 0.629 (0.606,0.662) | 0.651 (0.618,0.712) | 0.759 (0.815,0.778) | 0.662 (0.701,0.718) | 0.648 (0.763,0.72) | 0.469 (0.583,0.572) | 0.706 (0.705,0.715) | 0.7 (0.713,0.728) |
| 25 | 0.894 (0.871,0.923) | 0.894 (0.871,0.923) | 0.821 (0.895,0.783) | 0.702 (0.658,0.769) | 0.846 (0.902,0.82) | 0.388 (0.347,0.547) | 0.317 (0.527,0.27) | 0.532 (0.648,0.47) |
| 26 | 0.946 (0.933,0.961) | 0.946 (0.933,0.961) | 0.724 (0.737,0.713) | 0.647 (0.586,0.735) | 0.891 (0.953,0.859) | 0.646 (0.587,0.747) | 0.428 (0.725,0.351) | 0.951 (0.951,0.952) |
| 27 | 0.879 (0.928,0.842) | 0.781 (0.804,0.777) | 0.812 (0.886,0.751) | 0.928 (0.923,0.933) | 0.818 (0.89,0.772) | 0.933 (0.931,0.951) | 0.436 (0.786,0.425) | 0.222 (0.333,0.2) |
| 28 | 0.976 (0.968,0.984) | 0.698 (0.97,0.551) | 0.921 (0.985,0.866) | 0.983 (0.989,0.979) | 0.966 (0.988,0.947) | 0.978 (0.974,0.985) | 0.447 (0.984,0.296) | 0.47 (0.984,0.316) |
| 29 | 0.86 (0.932,0.8) | 0.949 (0.941,0.959) | 0.905 (0.962,0.854) | 0.949 (0.95,0.949) | 0.896 (0.953,0.852) | 0.389 (0.709,0.548) | 0.63 (0.861,0.602) | 0.739 (0.874,0.683) |
| 30 | 0.868 (0.921, 0.827) | 0.845 (0.898, 0.806) | 0.936 (0.965, 0.91) | 0.91 (0.903, 0.921) | 0.948 (0.959, 0.948) | 0.596 (0.532, 0.713) | 0.645 (0.877, 0.877) | 0.628 (0.864, 0.514) |

Table 27. GvHD

| flowMeans | flowMerge | SamSPECTRAL | ProClus |
|---|---|---|---|
| $F_1$ score (Precision, Recall) | | | |

| # | Completely automated | Partial Automated | Completely automated | Partial Automated | Completely automated | Partial Automated | Completely automated | Partial Automated |
|---|---|---|---|---|---|---|---|---|
| 1 | 0.801 (0.821,0.801) | 0.718 (0.65,0.804) | 0.496 (0.718,0.397) | 0.41 (0.484,0.369) | 0.831 (0.823,0.877) | 0.545 (0.501,0.674) | 0.368 (0.673,0.295) | 0.45 (0.672,0.4) |
| 2 | 0.74 (0.85,0.657) | 0.772 (0.719,0.84) | 0.492 (0.804,0.366) | 0.437 (0.581,0.36) | 0.874 (0.893,0.871) | 0.672 (0.622,0.769) | 0.499 (0.779,0.388) | 0.719 (0.686,0.759) |
| 3 | 0.809 (0.814,0.805) | 0.792 (0.743,0.853) | 0.491 (0.939,0.339) | 0.346 (0.482,0.275) | 0.758 (0.751,0.773) | 0.689 (0.644,0.782) | 0.246 (0.719,0.173) | 0.561 (0.775,0.483) |
| 4 | 0.772 (0.758,0.795) | 0.781 (0.728,0.844) | 0.563 (0.788,0.455) | 0.566 (0.682,0.496) | 0.816 (0.843,0.849) | 0.45 (0.402,0.598) | 0.359 (0.534,0.3) | 0.421 (0.484,0.439) |
| 5 | 0.727 (0.914,0.608) | 0.92 (0.905,0.939) | 0.683 (0.927,0.554) | 0.631 (0.741,0.561) | 0.617 (0.938,0.475) | 0.791 (0.748,0.854) | 0.343 (0.851,0.246) | 0.509 (0.85,0.425) |
| 6 | 0.773 (0.814,0.736) | 0.857 (0.821,0.899) | 0.544 (0.94,0.393) | 0.562 (0.703,0.475) | 0.63 (0.839,0.508) | 0.773 (0.73,0.842) | 0.368 (0.795,0.271) | 0.535 (0.752,0.424) |
| 7 | 0.681 (0.652,0.713) | 0.406 (0.329,0.552) | 0.522 (0.687,0.475) | 0.414 (0.375,0.479) | 0.65 (0.659,0.669) | 0.309 (0.254,0.476) | 0.493 (0.675,0.415) | 0.576 (0.602,0.554) |
| 8 | 0.361 (0.411,0.328) | 0.529 (0.445,0.66) | 0.047 (0.053,0.057) | 0.502 (0.465,0.552) | 0.599 (0.692,0.598) | 0.485 (0.453,0.627) | 0.587 (0.687,0.535) | 0.662 (0.626,0.714) |
| 9 | 0.381 (0.404,0.519) | 0.315 (0.24,0.479) | 0.396 (0.593,0.323) | 0.311 (0.265,0.4) | 0.819 (0.853,0.81) | 0.272 (0.222,0.441) | 0.497 (0.812,0.392) | 0.73 (0.709,0.758) |
| 10 | 0.819 (0.914,0.795) | 0.693 (0.62,0.787) | 0.412 (0.935,0.284) | 0.274 (0.461,0.206) | 0.976 (0.989,0.974) | 0.977 (0.983,0.981) | 0.476 (0.889,0.36) | 0.836 (0.865,0.816) |
| 11 | 0.84 (0.977,0.738) | 0.972 (0.965,0.98) | 0.383 (0.981,0.246) | 0.386 (0.938,0.252) | 0.978 (0.99,0.975) | 0.971 (0.973,0.98) | 0.376 (0.925,0.252) | 0.791 (0.959,0.685) |
| 12 | 0.952 (0.951,0.953) | 0.951 (0.943,0.959) | 0.329 (0.943,0.215) | 0.332 (0.882,0.22) | 0.962 (0.987,0.958) | 0.825 (0.799,0.877) | 0.401 (0.925,0.313) | 0.52 (0.882,0.397) |

Table 28. HSCT

| | F$_1$ score (Precision, Recall) | | | | | | | |
|---|---|---|---|---|---|---|---|---|
| | flowMeans | | flowMerge | | SamSPECTRAL | | ProClus | |
| # | Completely automated | Partial Automated | Completely automated | Partial Automated | Completely automated | Partial Automated | Completely automated | Partial Automated |
| 1 | 0.936 (0.969,0.906) | 0.732 (0.695,0.797) | 0.478 (0.957,0.34) | 0.303 (0.443,0.242) | 0.942 (0.973,0.918) | 0.385 (0.332,0.541) | 0.502 (0.669,0.435) | 0.579 (0.718,0.532) |
| 2 | 0.926 (0.956,0.899) | 0.947 (0.93,0.964) | 0.481 (0.977,0.324) | 0.464 (0.84,0.325) | 0.963 (0.991,0.941) | 0.833 (0.801,0.882) | 0.326 (0.834,0.214) | 0.635 (0.763,0.547) |
| 3 | 0.863 (0.917,0.814) | 0.974 (0.967,0.982) | 0.543 (0.959,0.387) | 0.543 (0.939,0.39) | 0.936 (0.975,0.918) | 0.963 (0.971,0.958) | 0.289 (0.921,0.184) | 0.655 (0.936,0.524) |
| 4 | 0.935 (0.956,0.916) | 0.97 (0.963,0.977) | 0.498 (0.957,0.345) | 0.498 (0.957,0.345) | 0.943 (0.963,0.933) | 0.888 (0.866,0.915) | 0.343 (0.916,0.224) | 0.674 (0.895,0.558) |
| 5 | 0.931 (0.988,0.882) | 0.959 (0.986,0.934) | 0.599 (0.993,0.436) | 0.604 (0.989,0.441) | 0.947 (0.994,0.905) | 0.987 (0.99,0.985) | 0.474 (0.717,0.363) | 0.601 (0.925,0.494) |
| 6 | 0.948 (0.951,0.945) | 0.855 (0.813,0.901) | 0.61 (0.956,0.46) | 0.557 (0.816,0.436) | 0.946 (0.971,0.923) | 0.864 (0.843,0.894) | 0.448 (0.812,0.389) | 0.786 (0.767,0.811) |
| 7 | 0.897 (0.971,0.836) | 0.973 (0.971,0.98) | 0.428 (0.985,0.277) | 0.475 (0.981,0.318) | 0.949 (0.991,0.914) | 0.326 (0.249,0.491) | 0.487 (0.747,0.364) | 0.573 (0.846,0.436) |
| 8 | 0.904 (0.933,0.878) | 0.795 (0.744,0.858) | 0.545 (0.928,0.407) | 0.463 (0.737,0.361) | 0.884 (0.951,0.831) | 0.381 (0.299,0.538) | 0.509 (0.64,0.441) | 0.287 (0.34,0.266) |
| 9 | 0.842 (0.982,0.763) | 0.384 (0.297,0.544) | 0.316 (0.461,0.247) | 0.25 (0.326,0.21) | 0.96 (0.99,0.933) | 0.976 (0.983,0.971) | 0.484 (0.694,0.408) | 0.352 (0.374,0.413) |
| 10 | 0.928 (0.939,0.918) | 0.568 (0.52,0.689) | 0.466 (0.993,0.308) | 0.342 (0.546,0.25) | 0.936 (0.994,0.89) | 0.986 (0.994,0.98) | 0.396 (0.657,0.322) | 0.408 (0.421,0.397) |
| 11 | 0.901 (0.982,0.833) | 0.958 (0.956,0.962) | 0.51 (0.978,0.358) | 0.534 (0.975,0.381) | 0.882 (0.983,0.803) | 0.568 (0.482,0.691) | 0.368 (0.699,0.272) | 0.582 (0.841,0.483) |
| 12 | 0.138 (0.15,0.129) | 0.885 (0.864,0.909) | 0.249 (0.535,0.17) | 0.291 (0.411,0.235) | 0.952 (0.979,0.931) | 0.362 (0.391,0.513) | 0.344 (0.668,0.247) | 0.438 (0.548,0.407) |
| 13 | 0.883 (0.929,0.841) | 0.934 (0.947,0.923) | 0.49 (0.948,0.342) | 0.321 (0.707,0.245) | 0.921 (0.963,0.89) | 0.452 (0.378,0.596) | 0.517 (0.914,0.384) | 0.567 (0.878,0.464) |
| 14 | 0.853 (0.905,0.81) | 0.826 (0.853,0.872) | 0.536 (0.978,0.373) | 0.251 (0.543,0.23) | 0.941 (0.986,0.902) | 0.352 (0.393,0.509) | 0.565 (0.733,0.495) | 0.52 (0.582,0.5) |
| 15 | 0.936 (0.969,0.906) | 0.732 (0.695,0.797) | 0.478 (0.957,0.34) | 0.303 (0.443,0.242) | 0.942 (0.973,0.918) | 0.385 (0.332,0.541) | 0.502 (0.669,0.435) | 0.579 (0.718,0.532) |
| 16 | 0.863 (0.979,0.777) | 0.896 (0.906,0.923) | 0.483 (0.979,0.33) | 0.504 (0.977,0.352) | 0.954 (0.987,0.926) | 0.462 (0.4,0.605) | 0.333 (0.737,0.235) | 0.366 (0.779,0.26) |
| 17 | 0.884 (0.922,0.851) | 0.949 (0.943,0.959) | 0.382 (0.67,0.275) | 0.561 (0.969,0.405) | 0.911 (0.947,0.879) | 0.451 (0.422,0.581) | 0.538 (0.79,0.45) | 0.478 (0.739,0.398) |
| 18 | 0.942 (0.943,0.944) | 0.953 (0.94,0.967) | 0.493 (0.915,0.344) | 0.222 (0.259,0.201) | 0.933 (0.945,0.924) | 0.351 (0.71,0.501) | 0.444 (0.692,0.341) | 0.483 (0.652,0.418) |
| 19 | 0.938 (0.968,0.912) | 0.969 (0.965,0.974) | 0.51 (0.969,0.354) | 0.221 (0.657,0.213) | 0.921 (0.949,0.897) | 0.395 (0.378,0.55) | 0.503 (0.829,0.367) | 0.489 (0.672,0.395) |
| 20 | 0.737 (0.914,0.64) | 0.956 (0.95,0.965) | 0.475 (0.967,0.325) | 0.229 (0.562,0.21) | 0.929 (0.979,0.894) | 0.441 (0.364,0.58) | 0.484 (0.628,0.414) | 0.575 (0.683,0.548) |
| 21 | 0.935 (0.963,0.915) | 0.954 (0.946,0.963) | 0.514 (0.955,0.374) | 0.532 (0.954,0.389) | 0.935 (0.973,0.903) | 0.337 (0.298,0.5) | 0.384 (0.659,0.285) | 0.592 (0.668,0.622) |
| 22 | 0.938 (0.945,0.933) | 0.601 (0.527,0.714) | 0.622 (0.946,0.476) | 0.418 (0.532,0.369) | 0.934 (0.967,0.908) | 0.494 (0.42,0.63) | 0.399 (0.588,0.321) | 0.355 (0.422,0.343) |
| 23 | 0.899 (0.93,0.87) | 0.899 (0.93,0.87) | 0.525 (0.978,0.368) | 0.525 (0.978,0.368) | 0.968 (0.983,0.956) | 0.386 (0.307,0.535) | 0.51 (0.734,0.405) | 0.441 (0.776,0.336) |

| 24 | 0.789 | 0.797 | 0.513 | 0.322 | 0.934 | 0.366 | 0.48 | 0.361 |
|  | (0.827,0.755) | (0.764,0.836) | (0.971,0.358) | (0.421,0.27) | (0.983,0.895) | (0.304,0.523) | (0.581,0.479) | (0.636,0.262) |
| 25 | 0.874 | 0.963 | 0.499 | 0.25 | 0.904 | 0.395 | 0.587 | 0.134 |
|  | (0.97,0.802) | (0.953,0.973) | (0.95,0.35) | (0.3,0.226) | (0.916,0.894) | (0.489,0.541) | (0.687,0.517) | (0.526,0.096) |
| 26 | 0.965 | 0.394 | 0.513 | 0.287 | 0.954 | 0.392 | 0.438 | 0.272 |
|  | (0.969,0.965) | (0.317,0.54) | (0.967,0.364) | (0.359,0.252) | (0.973,0.94) | (0.343,0.538) | (0.647,0.36) | (0.443,0.214) |
| 27 | 0.789 | 0.841 | 0.447 | 0.448 | 0.819 | 0.541 | 0.395 | 0.447 |
|  | (0.804,0.776) | (0.798,0.891) | (0.813,0.372) | (0.814,0.373) | (0.85,0.841) | (0.533,0.669) | (0.65,0.351) | (0.633,0.366) |
| 28 | 0.989 | 0.71 | 0.652 | 0.652 | 0.961 | 0.975 | 0.436 | 0.176 |
|  | (0.987,0.992) | (0.639,0.799) | (0.983,0.493) | (0.983,0.493) | (0.99,0.934) | (0.973,0.977) | (0.634,0.334) | (0.634,0.107) |
| 29 | 0.754 | 0.691 | 0.613 | 0.529 | 0.822 | 0.758 | 0.373 | 0.384 |
|  | (0.692,0.829) | (0.686,0.7) | (0.785,0.539) | (0.791,0.437) | (0.859,0.843) | (0.752,0.826) | (0.704,0.3) | (0.746,0.31) |
| 30 | 0.956 | 0.97 | 0.54 | 0.256 | 0.982 | 0.994 | 0.472 | 0.465 |
|  | (0.994,0.921) | (0.98,0.96) | (0.982,0.375) | (0.982,0.15) | (0.995,0.972) | (0.995,0.992) | (0.983,0.319) | (0.982,0.31) |

Table 29. WNV

| | F$_1$ score (Precision, Recall) | | | | | | | |
|---|---|---|---|---|---|---|---|---|
| | flowMeans | | flowMerge | | SamSPECTRAL | | ProClus | |
| # | Completely automated | Partial Automated | Completely automated | Partial Automated | Completely automated | Partial Automated | Completely automated | Partial Automated |
| 1 | 0.793 | 0.589 | 0.741 | 0.7 | 0.703 | 0.614 | 0.561 | 0.683 |
|  | (0.839,0.764) | (0.544,0.642) | (0.886,0.658) | (0.649,0.781) | (0.807,0.654) | (0.577,0.728) | (0.764,0.485) | (0.637,0.737) |
| 2 | 0.623 | 0.733 | 0.747 | 0.576 | 0.798 | 0.564 | 0.351 | 0.448 |
|  | (0.57,0.686) | (0.667,0.812) | (0.947,0.641) | (0.499,0.688) | (0.889,0.729) | (0.481,0.69) | (0.69,0.26) | (0.7,0.401) |
| 3 | 0.887 | 0.852 | 0.851 | 0.663 | 0.834 | 0.66 | 0.47 | 0.562 |
|  | (0.947,0.835) | (0.869,0.859) | (0.953,0.77) | (0.591,0.755) | (0.892,0.788) | (0.587,0.762) | (0.678,0.37) | (0.694,0.488) |
| 4 | 0.662 | 0.739 | 0.747 | 0.528 | 0.669 | 0.453 | 0.568 | 0.498 |
|  | (0.687,0.642) | (0.723,0.785) | (0.905,0.65) | (0.469,0.64) | (0.655,0.706) | (0.402,0.601) | (0.702,0.493) | (0.594,0.463) |
| 5 | 0.727 | 0.703 | 0.721 | 0.551 | 0.209 | 0.502 | 0.401 | 0.52 |
|  | (0.786,0.678) | (0.711,0.702) | (0.876,0.628) | (0.494,0.656) | (0.452,0.26) | (0.454,0.641) | (0.725,0.333) | (0.636,0.463) |
| 6 | 0.818 | 0.511 | 0.757 | 0.517 | 0.778 | 0.507 | 0.377 | 0.461 |
|  | (0.873,0.774) | (0.423,0.646) | (0.894,0.671) | (0.441,0.642) | (0.844,0.74) | (0.437,0.645) | (0.591,0.301) | (0.451,0.47) |
| 7 | 0.848 | 0.822 | 0.809 | 0.595 | 0.697 | 0.359 | 0.29 | 0.334 |
|  | (0.91,0.804) | (0.796,0.856) | (0.921,0.725) | (0.52,0.7) | (0.706,0.734) | (0.279,0.522) | (0.565,0.219) | (0.366,0.319) |
| 8 | 0.861 | 0.805 | 0.793 | 0.648 | 0.8 | 0.623 | 0.352 | 0.439 |
|  | (0.916,0.818) | (0.812,0.822) | (0.948,0.689) | (0.579,0.741) | (0.895,0.731) | (0.554,0.734) | (0.708,0.253) | (0.519,0.394) |
| 9 | 0.569 | 0.737 | 0.733 | 0.582 | 0.621 | 0.501 | 0.251 | 0.596 |
|  | (0.617,0.528) | (0.734,0.77) | (0.86,0.64) | (0.503,0.693) | (0.612,0.653) | (0.419,0.641) | (0.437,0.195) | (0.586,0.632) |
| 10 | 0.745 | 0.423 | 0.766 | 0.445 | 0.622 | 0.409 | 0.446 | 0.467 |
|  | (0.77,0.743) | (0.355,0.529) | (0.881,0.688) | (0.468,0.574) | (0.617,0.657) | (0.343,0.565) | (0.622,0.399) | (0.414,0.54) |
| 11 | 0.845 | 0.82 | 0.784 | 0.637 | 0.825 | 0.614 | 0.534 | 0.521 |
|  | (0.877,0.815) | (0.842,0.799) | (0.907,0.697) | (0.573,0.737) | (0.897,0.786) | (0.559,0.727) | (0.687,0.461) | (0.764,0.438) |
| 12 | 0.818 | 0.852 | 0.27 | 0.703 | 0.866 | 0.702 | 0.428 | 0.413 |
|  | (0.954,0.719) | (0.927,0.796) | (0.951,0.18) | (0.64,0.782) | (0.948,0.804) | (0.709,0.792) | (0.758,0.315) | (0.798,0.297) |
| 13 | 0.696 | 0.695 | 0.713 | 0.714 | 0.71 | 0.547 | 0.37 | 0.318 |
|  | (0.802,0.64) | (0.795,0.65) | (0.887,0.62) | (0.751,0.743) | (0.899,0.628) | (0.517,0.676) | (0.715,0.301) | (0.462,0.281) |

# Appendices 2: Details of Real Datasets

Table 30. DLBCL

| Samples | # Events | # Cluster 1 | # Cluster 2 | # Cluster 3 | # Cluster 4 |
|---|---|---|---|---|---|
| 1 | 3796 | 1561 | 2209 | | |
| 2 | 7396 | 6860 | 477 | | |
| 3 | 5524 | 604 | 4873 | | |
| 4 | 3290 | 996 | 274 | 1912 | |
| 5 | 4284 | 3028 | 969 | 131 | |
| 6 | 12369 | 2011 | 10333 | | |
| 7 | 2886 | 147 | 966 | 1423 | |
| 8 | 7985 | 5337 | 2593 | | |
| 9 | 3569 | 1291 | 2156 | | |
| 10 | 1856 | 346 | 1417 | | |
| 11 | 4077 | 1127 | 2195 | 272 | |
| 12 | 8149 | 1319 | 1159 | 4595 | 946 |
| 13 | 8183 | 1561 | 1494 | 62 | 4815 |
| 14 | 18243 | 139 | 5767 | 11938 | |
| 15 | 24564 | 16807 | 1098 | 5827 | |
| 16 | 9513 | 175 | 1132 | 8165 | |
| 17 | 13393 | 3963 | 1436 | 1284 | 4746 |
| 18 | 12621 | 9818 | 826 | 1478 | |
| 19 | 10197 | 8591 | 157 | 1210 | |
| 20 | 12563 | 4095 | 7363 | 804 | |
| 21 | 10000 | 1646 | 1380 | 6686 | |
| 22 | 11908 | 798 | 797 | 10174 | |
| 23 | 8835 | 335 | 4434 | 794 | 3078 |
| 24 | 17257 | 6386 | 3630 | 1342 | 2620 |
| 25 | 11729 | 6410 | 1702 | 2861 | |
| 26 | 16683 | 8904 | 3672 | 3603 | |
| 27 | 14290 | 8252 | 1074 | 4366 | |
| 28 | 21011 | 20674 | 183 | | |
| 29 | 10715 | 261 | 4318 | 5879 | |
| 30 | 11792 | 375 | 245 | 2443 | 8396 |

Table 31. GvHD

| Samples | # Events | # Cluster 1 | # Cluster 2 | # Cluster 3 | # Cluster 4 | # Cluster 5 |
|---|---|---|---|---|---|---|
| 1 | 13831 | 1846 | 784 | 1040 | 9303 | |
| 2 | 14936 | 1082 | 780 | 789 | 11479 | |
| 3 | 15227 | 11886 | 1162 | 1133 | 649 | |
| 4 | 12751 | 2145 | 7622 | 1171 | 1249 | |
| 5 | 17008 | 415 | 1220 | 14502 | 310 | |
| 6 | 14806 | 1044 | 12459 | 930 | | |
| 7 | 13773 | 3605 | 6568 | 733 | 880 | 586 |
| 8 | 14787 | 3014 | 9259 | 608 | 1034 | |
| 9 | 17640 | 712 | 209 | 783 | 7028 | 7751 |
| 10 | 23377 | 17467 | 929 | 4528 | | |
| 11 | 32700 | 873 | 31174 | 67 | | |
| 12 | 16336 | 1402 | 14300 | 86 | | |

Table 32. HSCT

| Samples | # Events | # Cluster 1 | # Cluster 2 | # Cluster 3 | # Cluster 4 |
|---|---|---|---|---|---|
| 1 | 9936 | 14 | 397 | 8621 | 754 |
| 2 | 9655 | 8515 | 284 | 800 | |
| 3 | 9790 | 9135 | 505 | 47 | |
| 4 | 9694 | 570 | 8856 | 128 | 37 |
| 5 | 9638 | 1803 | 7767 | | |
| 6 | 9783 | 7302 | 772 | 1518 | |
| 7 | 9968 | 4907 | 4822 | 175 | |
| 8 | 9179 | 1001 | 4939 | 2947 | |
| 9 | 9780 | 4271 | 3461 | 906 | 1064 |
| 10 | 9055 | 484 | 6178 | 2359 | |
| 11 | 8914 | 6154 | 1671 | 989 | |
| 12 | 9928 | 4301 | 1000 | 3968 | 513 |
| 13 | 9896 | 342 | 5911 | 3414 | |
| 14 | 9206 | 4709 | 3313 | 1147 | |
| 15 | 9155 | 1574 | 2476 | 4942 | |
| 16 | 9073 | 5491 | 2863 | 638 | |
| 17 | 9517 | 5518 | 2988 | 592 | 320 |
| 18 | 9659 | 4720 | 101 | 4574 | 41 |
| 19 | 9829 | 324 | 4013 | 5398 | |
| 20 | 9554 | 539 | 3769 | 5090 | |
| 21 | 8390 | 4180 | 2429 | 1619 | |
| 22 | 9677 | 5123 | 971 | 828 | 2505 |
| 23 | 9181 | 273 | 4077 | 4770 | |
| 24 | 9353 | 3625 | 698 | 4222 | 708 |
| 25 | 8931 | 480 | 3853 | 4412 | |
| 26 | 6235 | 374 | 2401 | 2741 | 627 |
| 27 | 9399 | 2024 | 88 | 6264 | |
| 28 | 9536 | 79 | 7549 | 1845 | |
| 29 | 7210 | 5906 | 69 | 361 | 21 |
| 30 | 8885 | 8724 | 136 | | |

Table 33. WNV

| Samples | # Events | # Cluster 1 | # Cluster 2 | # Cluster 3 | # Cluster 4 |
|---|---|---|---|---|---|
| 1 | 92498 | 5586 | 543 | 11861 | 67265 |
| 2 | 92607 | 63845 | 12146 | 2470 | 12142 |
| 3 | 91738 | 69862 | 18278 | 734 | 769 |
| 4 | 95660 | 6205 | 7875 | 17525 | 57505 |
| 5 | 91674 | 1702 | 4063 | 58719 | 17939 |
| 6 | 92426 | 658 | 59603 | 22550 | 2580 |
| 7 | 94749 | 4457 | 18398 | 49493 | 19087 |
| 8 | 87174 | 937 | 2962 | 64002 | 16897 |
| 9 | 94647 | 60631 | 22454 | 2198 | 6659 |
| 10 | 93264 | 15604 | 5154 | 12542 | 52708 |
| 11 | 94489 | 1417 | 337 | 68690 | 16932 |
| 12 | 90036 | 16517 | 71294 | 413 | |
| 13 | 103411 | 69881 | 20148 | 943 | 1437 |